%% file: root.tex
\documentclass[letterpaper, 10 pt, journal, twoside]{IEEEtran}

\usepackage{graphicx}
\usepackage{setspace}
\usepackage[T1]{fontenc}
\usepackage{multicol}
\usepackage{amsmath,amssymb}
\usepackage{epstopdf}
\usepackage{graphicx,color}
\usepackage{xspace}
\usepackage{balance}
\usepackage[per-mode=symbol,binary-units=true]{siunitx}
\usepackage{ textcomp } 
\usepackage{booktabs}
\usepackage{paralist}
\usepackage{layouts}
\usepackage{url}
\usepackage{bm}
\usepackage{tikz}
\usepackage[percent]{overpic}
\usepackage{tabularx} 
\usetikzlibrary{fit,shapes.callouts,shapes.geometric,backgrounds,positioning,arrows.meta}

\newcommand{\ie}{i.e.,\ }
\newcommand{\eg}{e.g.,\ }

\newlength{\figureheight}
\newcommand{\etal}{\xspace{}et al.\xspace}

\newcommand{\reffig}[1]{Fig.~\ref{#1}}
\newcommand{\reftab}[1]{Tab.~\ref{#1}}
\newcommand{\refsec}[1]{Sec.~\ref{#1}}

\begin{document}

\title{Fast Autonomous Flight in Warehouses for Inventory Applications}

\author{Marius Beul, David Droeschel, Matthias Nieuwenhuisen, Jan Quenzel, Sebastian Houben, and Sven Behnke%
\thanks{Manuscript received: February 23, 2018; Revised: May 17, 2018; Accepted: June 6, 2018.}
\thanks{This paper was recommended for publication by Editor Jonathan Roberts upon evaluation of the Associate Editor and Reviewers' comments.

This work was supported by the German Bundesministerium f\"ur Wirtschaft und Energie in the Autonomics for Industry 4.0 project InventAIRy, and grants BE 2556/7-2 and BE 2556/8-2 of the German Research Foundation (DFG).}%
\thanks{The authors are with the Autonomous Intelligent Systems Group, University of Bonn, Germany
        {\tt\footnotesize mbeul@ais.uni-bonn.de}}%
\thanks{Digital Object Identifier (DOI): see top of this page.}
}

\markboth{IEEE Robotics and Automation Letters. Preprint Version. Accepted June, 2018}
{Beul \MakeLowercase{\textit{et al.}}: Fast Autonomous Flight in Warehouses}

\maketitle

\begin{abstract}
The past years have shown a remarkable growth in use-cases for micro aerial vehicles (MAVs). Conceivable indoor applications require highly robust environment perception, fast reaction to changing situations, and stable navigation, but reliable sources of absolute positioning like GNSS or compass measurements are unavailable during indoor flights.

We present a high-performance autonomous inventory MAV for operation inside warehouses. The MAV navigates along warehouse aisles and detects the placed stock in the shelves alongside its path with a multimodal sensor setup containing an RFID reader and two high-resolution cameras. We describe in detail the SLAM pipeline based on a 3D lidar, the setup for stock recognition, the mission planning and trajectory generation, as well as a low-level routine for avoidance of dynamical or previously unobserved obstacles. Experiments were performed in an operative warehouse of a logistics provider, in which an external warehouse management system provided the MAV with high-level inspection missions that are executed fully autonomously.
\end{abstract}


\section{Introduction}
\label{sec:Introduction}
\input{introduction.tex}

\section{Related Work}
\label{sec:Related_Work}
\input{related_work.tex}

\section{System Setup}
\label{sec:System_Setup}
\input{system_setup.tex}

\section{Environment Perception}
\label{sec:Environment_Perception}
\input{environment_perception.tex}

\section{Navigation and Control}
\label{sec:Navigation_and_Control}
\input{navigation_and_control.tex}

\section{Evaluation}
\label{sec:Evaluation}
\input{evaluation.tex}

\section{Conclusion}
\label{sec:Conclusion}
\input{conclusion.tex}


\bibliographystyle{IEEEtran}
\bibliography{literature_references}

\end{document}

%% file: introduction.tex
\IEEEPARstart{I}{n} the last years, many novel applications for flying robots emerged, enabled by two main factors: i) manufacturers  developed affordable and capable micro aerial vehicles (MAVs) for hobby, recreation and professional usage that do not require extensive flight training; ii) recent advances in robotic research led to efficient methods for environment perception and safe navigation, enabling various applications that can only be performed autonomously.
This includes operations at high velocities and close to structures. Both conditions are prohibitive for safe operation by a human pilot.
One driver for developing such systems is also the DARPA-formulated goal of flying fast and autonomously in cluttered environments without GPS and external sensing or control in their Fast Lightweight Autonomy Program (FLA) \cite{darpaFLA}.

\begin{figure}[t]
  \centering
  \begin{tikzpicture}
      \definecolor{red}{rgb}{0.7,0.0,0.0}
      \node[inner sep = 0,anchor=north west] at (0,0) {\includegraphics[trim=010mm 000mm 000mm 070mm,clip,width=1.00\columnwidth]{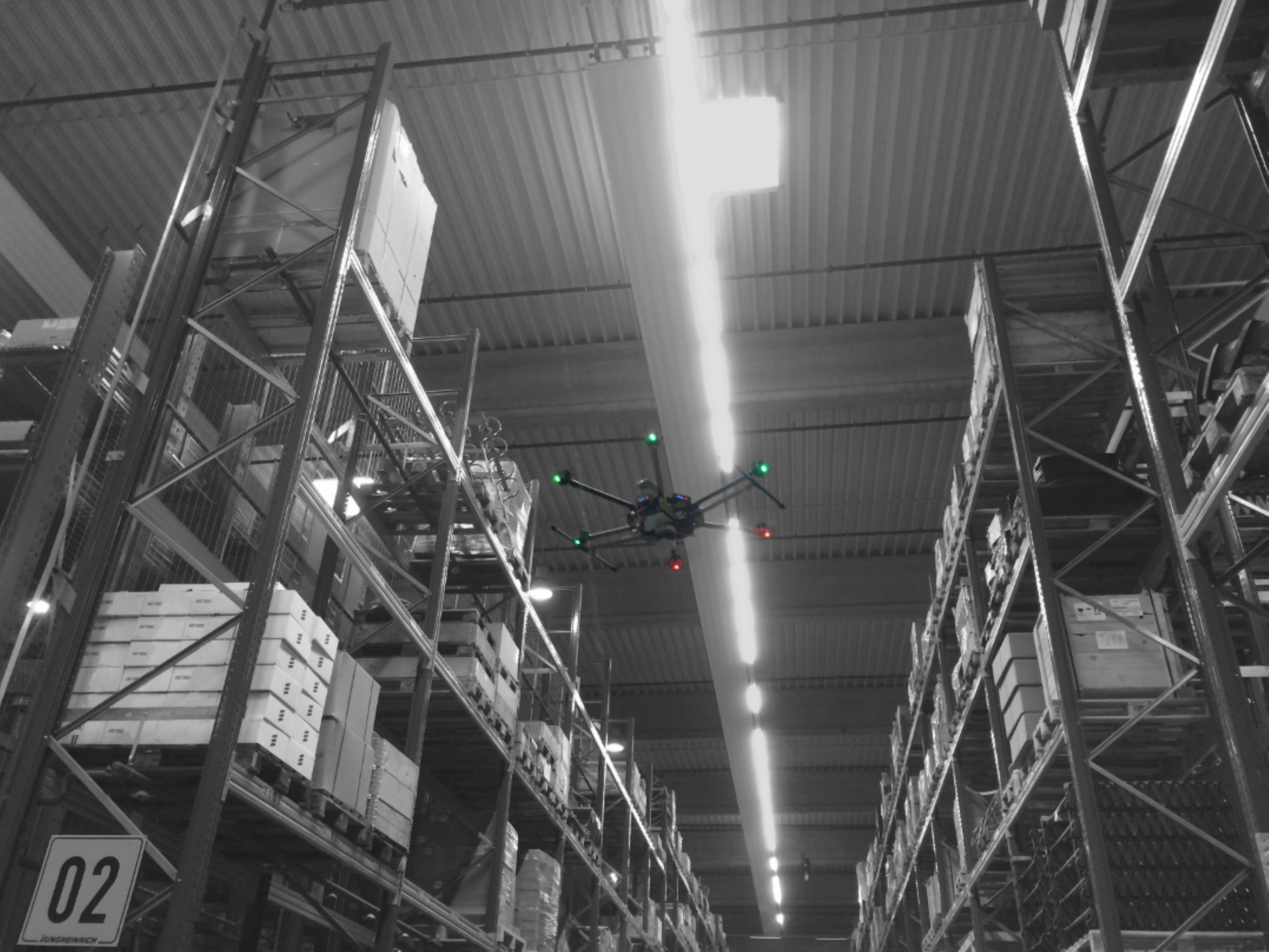}};
      \draw[line width=0.4mm, red] (4.5,-1.90) circle (1.2cm and 0.6cm);
  \end{tikzpicture}
  \vspace{-2.0em}
  \caption{Our inventory system performs a fully autonomous inspection of a warehouse. The main challenges are the fast navigation in narrow passages close to structures and the localization in a large self-similar indoor environment with distant walls.}
  \label{fig:inventur_screenshot}
  \vspace{-3ex}
\end{figure}

While in most current applications, MAVs maintain a safe distance from the object to inspect or follow; many future applications require the MAV to operate close to obstacles or even in restricted indoor spaces. As an example, in this paper, we consider the use case of automatic inventory in a warehouse.
It requires the MAV to quickly detect, identify, and map the stored items.
In this way, it is possible to keep an always-up-to-date inventory record of the contents within the warehouse. Current commercial systems~\cite{dronescan,eyesee} for this task merely deploy a scanner on the platform and perform a piloted flight in order to read tags on the goods.

Autonomous maneuvering inside such a building is highly challenging as most of the space is occupied with high shelves filled with stocked goods as shown in \reffig{fig:inventur_screenshot}.
This leaves only small aisles for navigation which might also be obstructed by other objects like forklifts. Additionally, the shelf rows lack distinctive geometric features and are highly self-similar which makes precise self-localization difficult.
On the other hand, these narrow structures are embedded in large halls with stable, but far-away localization aids like walls.
This requires real-time localization with long-distance sensors in large maps with many structures.

We present our self-localization and mapping approach based on a 3D lidar, which is able to handle these challenging situations robustly. The lidar is also the basis of a low-level obstacle avoidance mechanism. In addition, the robot carries a sensor setup to identify the stocked material by means of fiducial markers and RFID tags. The flight mission is provided by a warehouse management system (WMS) as a sequence of storage panels that have to be inspected. The mission is planned in a semantic, yet metric, map of the warehouse that contains the approximate placing of all the shelf rows and the number and relative position of the storage panels within. The laser-based map is aligned with this representation in order to define the inspection poses that the robot consecutively visits during its flight.

Experiments are performed in a warehouse of a logistics provider containing narrow aisles between shelves and larger open areas.
We mapped several shelf rows and performed autonomous inventory missions including the transition between rows and the avoidance of static obstacles.
Furthermore, we demonstrated the reactive avoidance of dynamic obstacles approaching the MAV.
We discuss guidelines for the development of future systems for autonomous indoor operation and draw prospects for the future of autonomous inventory robots.
To demonstrate the robustness of our localization and control at high velocities, not reachable in our indoor environments due to the required acceleration distance, we further evaluate our system outdoors with flights reaching velocities over \SI{28}{\kilo\meter\per\hour} without GNSS feedback.

In our integrated system, we employ and extend methods based on our own previous work: The SLAM system is detailed in~\cite{Droeschel2017104} and~\cite{Droeschel:ICRA2014}. Our obstacle avoidance extends~\cite{nieuwenhuisen2013isprs} and the mechanics of our model predictive controller (MPC) are described in~\cite{beul2017icuas}.

\noindent Our main contributions are
\begin{compactitem}
    \item robust self-localization solely based on an onboard lidar at high velocities up to \SI{7.8}{\meter\per\second} (Sec.~\ref{sec:Environment_Perception}),
    \item fast fully autonomous navigation and control, including avoidance of static and dynamic obstacles in indoor and outdoor environments (Sec.~\ref{sec:Navigation_and_Control}),
    \item an integrated autonomous robot system for aerial stocktaking with multimodal tag detection, evaluated in an operative warehouse (Sec.~\ref{sec:Evaluation}).
\end{compactitem}

%% file: related_work.tex
\begin{figure}[t]
  \centering
  \resizebox{0.9\linewidth}{!}{
  \input{MAV.pgf}
  }
  \vspace{-1.0em}
  \caption{Design of our MAV equipped with a Velodyne Puck LITE, fast onboard computer, two synchronized global shutter color cameras, and an RFID reader. The landing feet are retractable to allow for true \SI{360}{\degree} perception.}
  \label{fig:MAV}
  \vspace{-3ex}
\end{figure}
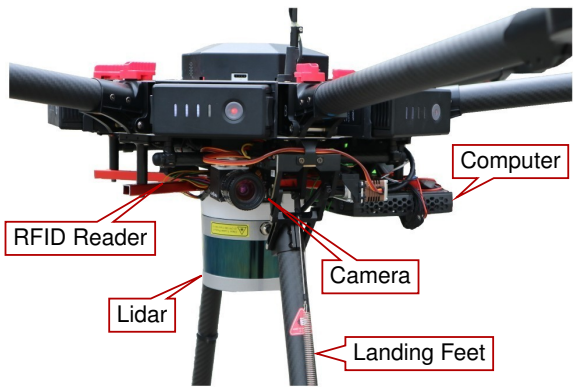

Today, fast MAV flight without external sensing is mostly vision-based.
Recently, Falanga \etal~\cite{falanga2017} presented an MAV flying with \SI{3}{\meter\per\second} through narrow gaps.
This requires precise relative localization and navigation. We focus on fast navigation in allocentric maps with reliable obstacle avoidance, which currently is not achievable by using cameras alone.

Shen \etal~\cite{Shen_RSS_2013} present an MAV that is capable of autonomous vision-based flight with up to \SI{4}{\meter\per\second} on a straight line, \SI{2}{\meter\per\second} on a figure eight, and \SI{1.5}{\meter\per\second} in an indoor environment. Although the system is relatively fast, the authors report significant drift, induced by solely relying on cameras for state estimation.

Another vision-based, lightweight MAV system has been presented by Burri \etal~\cite{burri2012}.
Their work focuses on industrial boiler inspection with agile flight in industrial environments.

Florence \etal~\cite{florence2016} use a combination of vision and a 2D laser scanner to avoid obstacles at high velocities. Their system flies in cluttered unknown environments with large state uncertainties.
For our application, we rely on precise, but still fast, allocentric localization and assume that an allocentric map contains the major, complex obstacles.

Our targeted scenario is particularly adverse for the use of visual perception. On one hand, panels and shelf rows carry highly similar and repetitive visual cues, which preclude any form of place recognition. On the other hand, local geometry is also highly self-similar and symmetric. Hence, we solely rely on high-frequency 3D laser scans for obstacle perception and state estimation. Its large field-of-view and long measurement range allow for resolving local similarities by using large-scale structures for localization.

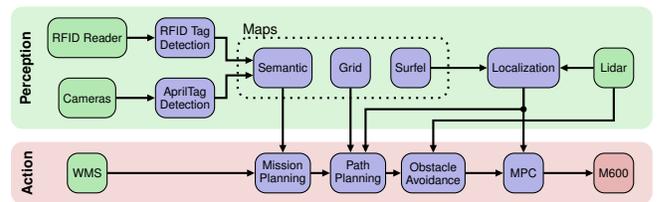
\begin{figure}[t]
  \centering
  \input{system_diagram.pgf}
  \vspace{-0.5em}
  \caption{System overview. Inputs are depicted in green and software components in blue. An external warehouse management system (WMS) provides an unordered list of waypoints of to be inspected goods to the mission planning. Control commands are sent to the SDK of the DJI Matrice 600 (red).}
  \label{fig:system_diagram}
  \vspace{-3ex}
\end{figure}

Ma \etal~\cite{ma2017drone} addressed automatic inventory with a lightweight Parrot drone.
Their RFly system relays the RFID signal to a reader and is able to triangulate the location of the tag with a reported accuracy of below \SI{20}{\centi\meter}.
However, in order to self-localize the robot, they rely on an external motion capturing setup, which limits the practical feasibility.

Similar to our system, Ortiz \etal~\cite{ortiz2014} perform inspection in narrow spaces.
They developed a quadrotor MAV for autonomous vessel inspection.
A combination of laser localization and visual odometry yields a 2D localization approach decoupled from the height measurements.
Our system performs SLAM with 6D pose estimation based on a high-performance 3D lidar.

An early version of our inventory MAV~\cite{ROS_book_2017} relied on a combination of two rotating 2D lidars for a low frequency localization and mapping with visual odometry performed with three pairs of wide-angle stereo cameras.
Although the system proved itself robust, the setup proposed in this paper only relies on a single 3D lidar as primary sensor and, hence, significantly reduces the overall system complexity.
Furthermore, the increased frequency of \SI{360}{\degree} scans obviates the requirement for additional visual odometry.
To account for the narrower vertical field of view in the proposed system, our path planning optionally limits the ascension and descension angle.

%% file: MAV.pgf
\begin{tikzpicture}[auto]
    \definecolor{red}{rgb}{0.7,0.0,0.0}
    \node[anchor=south west,inner sep=0] (image) at (0,0) {\includegraphics[width=1.0\linewidth]{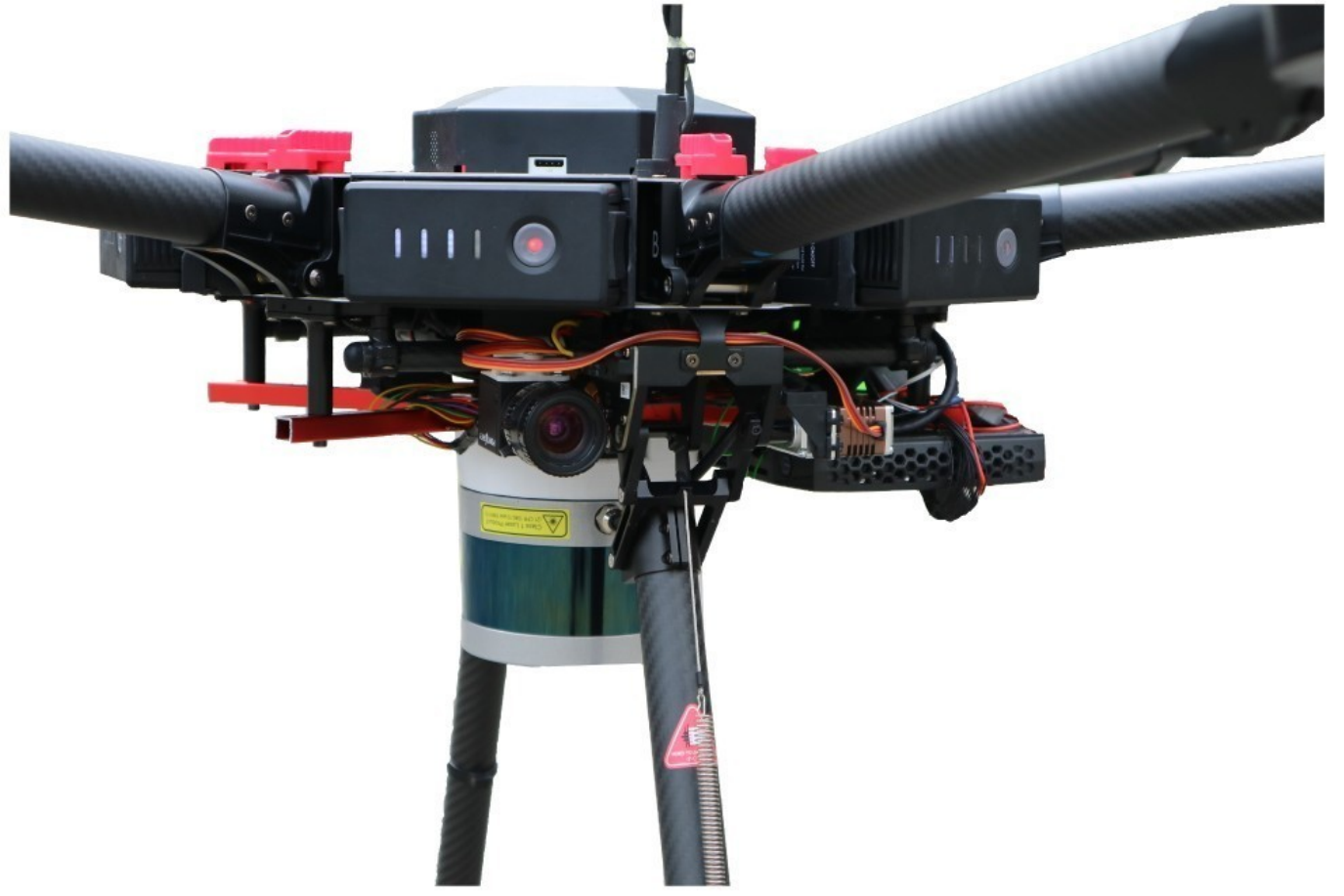}};
    \begin{scope}[x={(image.south east)},y={(image.north west)},font=\sffamily,every node/.style={align=center},line/.style={red, line width=1pt},box/.style={rectangle,draw=red,inner sep=0.3333em,thick},every node/.style={align=center,text height=1.5ex,text depth=.25ex,text centered},]
        \node[box,anchor=south east,rectangle callout,callout relative pointer={(-0.05,-0.05)}] (cam1) at (0.98, 0.55) {Computer};
        \node[box,anchor=south west,rectangle callout,callout relative pointer={( 0.06, 0.06)}] (cam2) at (0.18, 0.15) {Lidar};
        \node[box,anchor=south west,rectangle callout,callout relative pointer={(-0.14, 0.16)}] (cam2) at (0.55, 0.25) {Camera};
        \node[box,anchor=south west,rectangle callout,callout relative pointer={(-0.05, 0.00)}] (cam2) at (0.59, 0.05) {Landing Feet};
        \node[box,anchor=south west,rectangle callout,callout relative pointer={( 0.10, 0.10)}] (cam2) at (0.00, 0.35) {RFID Reader};
    \end{scope}
\end{tikzpicture}

%% file: system_diagram.pgf
\begin{tikzpicture}[font=\sffamily\small,>={Stealth[inset=0pt,length=3pt,angle'=45]}]

\tikzset{sensor_node/.append style={minimum size=1.0em,minimum height=3.0em,minimum width=3.0em,draw,align=center,rounded corners,scale=0.5,fill=green!30!white}}
\tikzset{actor_node/.append style={minimum size=1.0em,minimum height=3.0em,minimum width=3.0em,draw,align=center,rounded corners,scale=0.5,fill=red!30!white}}
\tikzset{content_node/.append style={minimum size=1.0em,minimum height=3.0em,minimum width=3.0em,draw,align=center,rounded corners,scale=0.5,fill=blue!30!white}}
\tikzset{label_node/.append style={scale=0.6}}


\definecolor{red}{rgb}     {0.7,0.0,0.0}
\definecolor{green}{rgb}   {0.0,0.7,0.0}
\definecolor{blue}{rgb}    {0.0,0.0,0.7}
\definecolor{grey}{rgb}    {0.5,0.5,0.5}

\draw[thick, rounded corners, green!15!white,fill] (-3.8,2.2) -- (4.7,2.2) -- (4.7,0.6) -- (-3.8,0.6) -- cycle;
\draw[thick, rounded corners, red!15!white,fill] (-3.8,0.4) -- (4.7,0.4) -- (4.7,-0.4) -- (-3.8,-0.4) -- cycle;
\draw[thick, rounded corners, dotted] (-0.8,1.8) -- (2.0,1.8) -- (2.0,1.0) -- (-0.8,1.0) -- cycle;

\node(explore_label)[label_node] at (-3.6,1.4){\rotatebox{90}{\textbf{Perception}}};
\node(explore_label)[label_node] at (-3.6,0.0){\rotatebox{90}{\textbf{Action}}};
\node(explore_label)[label_node] at (-0.5,1.9){Maps};

\node(RFID_Reader)[sensor_node] at (-2.8,1.8) {RFID Reader};
\node(Cameras)[sensor_node] at (-2.8,1.0) {Cameras};
\node(WMS)[sensor_node] at (-2.8,0.0) {WMS};
\node(Lidar)[sensor_node] at (4.2,1.4) {Lidar};
\node(M600)[actor_node] at (4.2,0.0) {M600};

\node(RFID_Tag_Detection)[content_node] at (-1.5,1.8) {RFID Tag\\Detection};
\node(AprilTag_Detection)[content_node] at (-1.5,1.0) {AprilTag\\Detection};
\node(Semantic)[content_node] at (-0.2,1.4) {Semantic};
\node(Grid)[content_node] at (0.7,1.4) {Grid};
\node(Surfel)[content_node] at (1.5,1.4) {Surfel};

\node(Localization)[content_node] at (3.0,1.4) {Localization};
\node(Mission_Planning)[content_node] at (-0.2,0.0) {Mission\\Planning};
\node(Path_Planning)[content_node] at (0.8,0.0) {Path\\Planning};
\node(Obstacle_Avoidance)[content_node] at (1.8,0.0) {Obstacle\\Avoidance};
\node(MPC)[content_node] at (3.0,0.0) {MPC};

\draw[->, thick] (Lidar) -- (Localization);
\draw[->, thick] (Surfel) -- (Localization);
\draw[->, thick] (Cameras) -- (AprilTag_Detection);
\draw[->, thick] (RFID_Reader) -- (RFID_Tag_Detection);
\draw[->, thick] (WMS) -- (Mission_Planning);
\draw[->, thick] (Localization) -- (MPC);
\draw[->, thick] (Semantic) -- (Mission_Planning);
\draw[->, thick] (Mission_Planning) -- (Path_Planning);
\draw[->, thick] (Path_Planning) -- (Obstacle_Avoidance);
\draw[->, thick] (Obstacle_Avoidance) -- (MPC);
\draw[->, thick] (MPC) -- (M600);

\draw[->, thick] (Localization) -- (3.0,0.85) -- (0.9,0.85) -- (0.9,0.27);
\draw[->, thick] (Grid) -- (0.7,0.27);
\draw[->, thick] (Lidar) -- (4.2,0.7) -- (1.8,0.7) -- (Obstacle_Avoidance);

\draw[->, thick] (RFID_Tag_Detection) -- (-0.95,1.8) -- (-0.95,1.5) -- (-0.59,1.5);
\draw[->, thick] (AprilTag_Detection) -- (-0.95,1.0) -- (-0.95,1.3) -- (-0.59,1.3);

\fill (3.0,0.85) circle [radius=1.2pt];

\end{tikzpicture}

%% file: system_setup.tex
\begin{figure*}[t]
  \centering
  \includegraphics[trim=20mm 00mm 10mm 20mm,clip,height=0.23\textwidth]{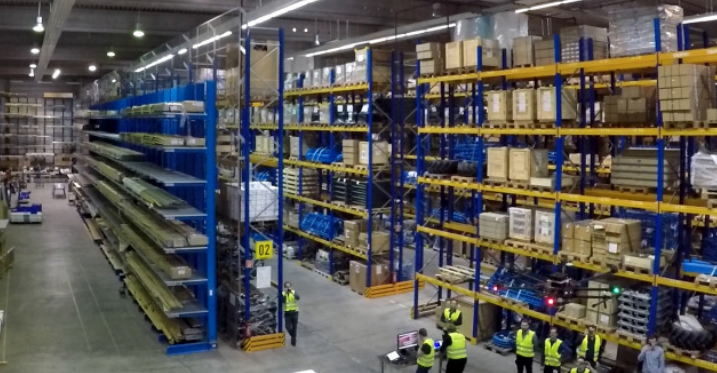}~ %
  \includegraphics[trim=00mm 00mm 00mm 45mm,clip,height=0.23\textwidth]{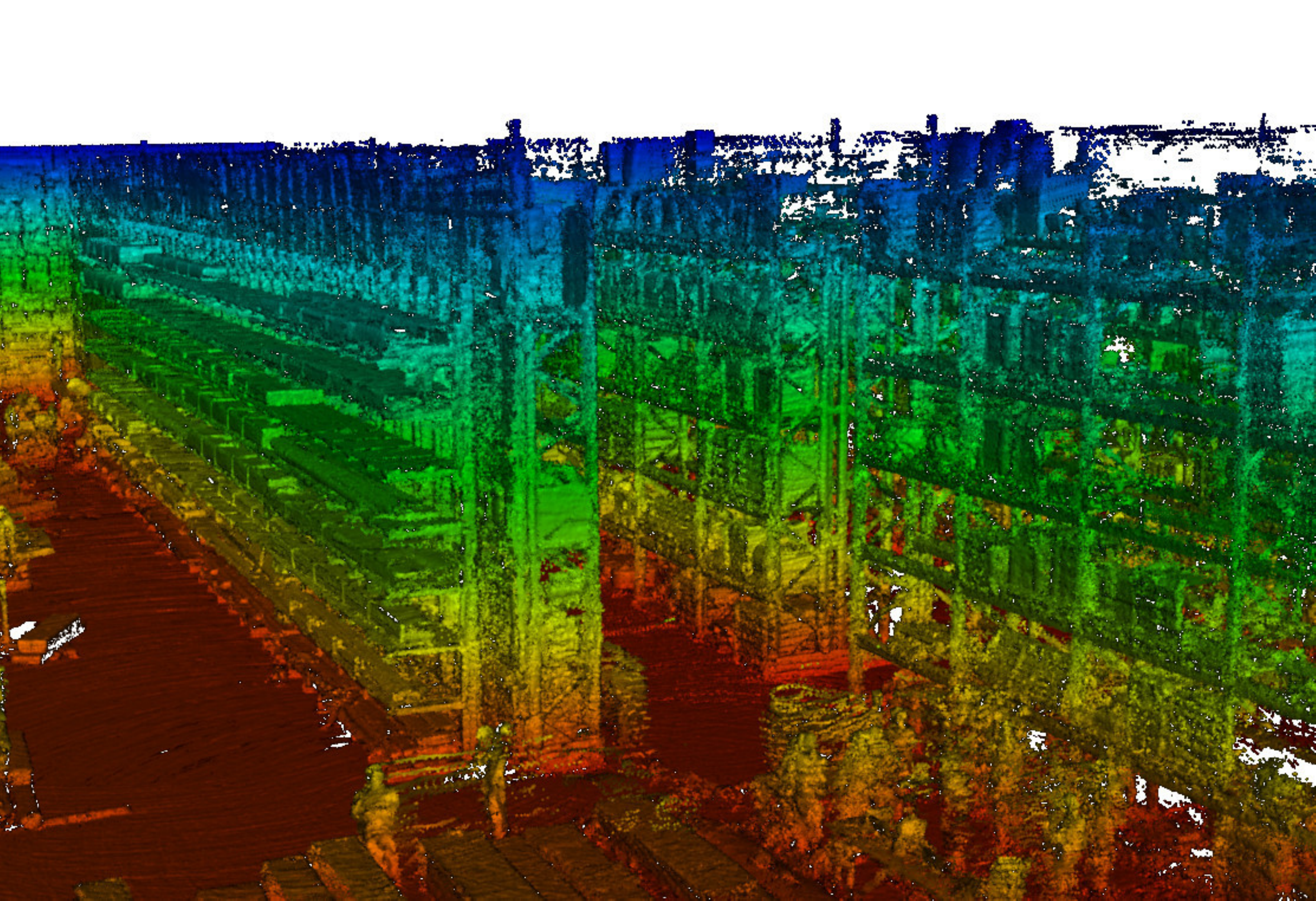}~
    \begin{tikzpicture}
        \definecolor{green}{rgb}   {0.0,0.7,0.0}
        \node[inner sep = 0,anchor=north west] at (0,0) {\includegraphics[trim=05mm 09mm 00mm 10mm,clip,height=0.23\textwidth]{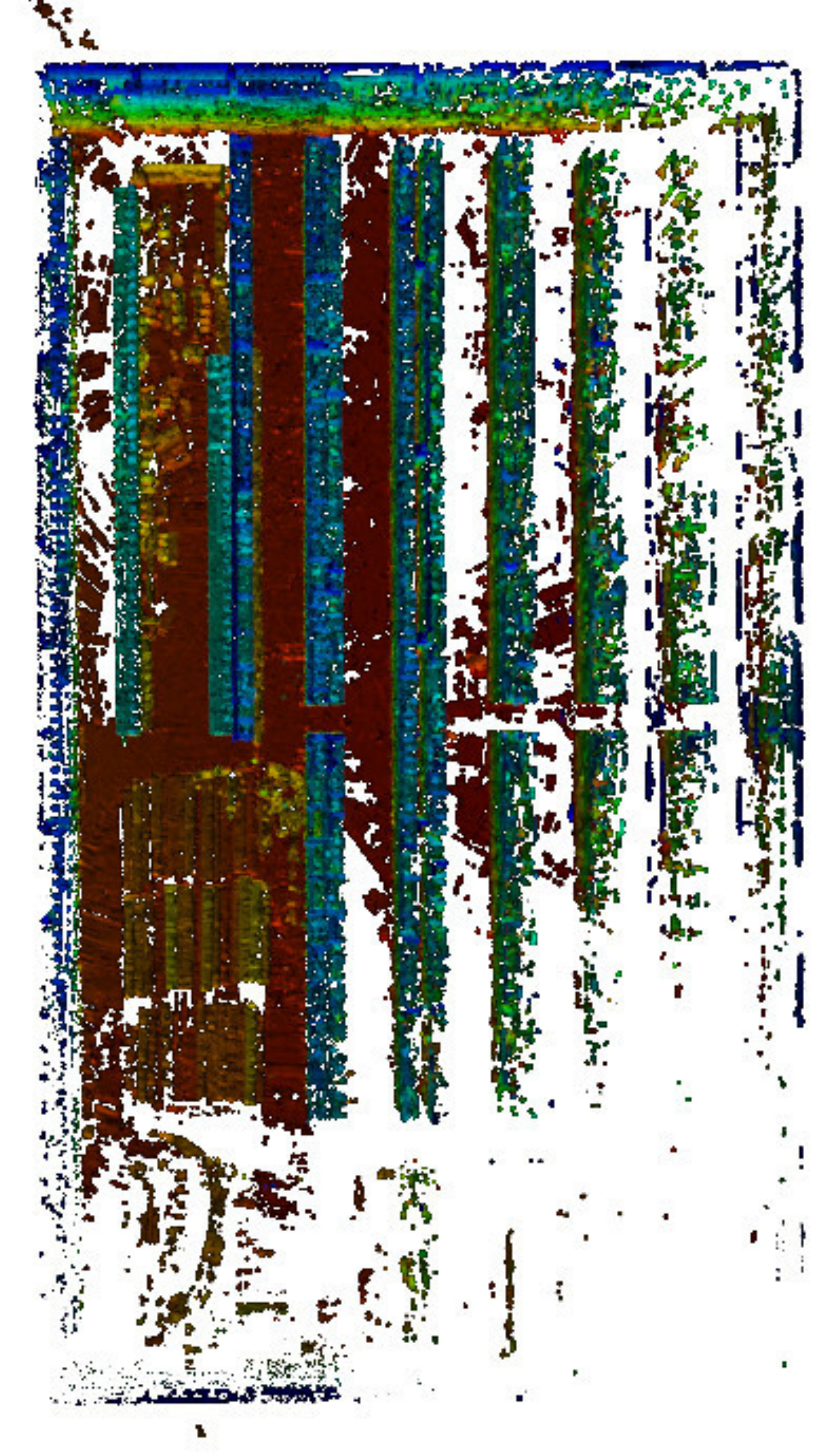}};
        \draw[thick,fill=green,draw=black,fill opacity=0.5] (0.4,-3.0) -- (0.9,-2.8) arc (24:88:0.54) -- cycle;
        \fill (0.4,-3.0) circle [radius=2pt];
  \end{tikzpicture}
  \vspace{-4.5ex}
  \caption{3D map from the initial manual flight. The top-down view (right) shows the dimensions of the acquired map of the \SI[product-units = single]{100 x 60}{\meter} warehouse. The camera perspective is highlighted in green. The warehouse contains tight, self-repetitive, and cluttered structures like shelves and stock, and larger, far-away structures like walls. For robust localization, the MAV has to employ a map of the structure of the large building.}
  \label{fig:mapping}
  \vspace{-3ex}
\end{figure*}

Our MAV, shown in \reffig{fig:MAV}, is based on the DJI Matrice~600 platform with a diameter of approximately $\SI{170}{\centi\meter}$. It is equipped with a lightweight, yet powerful, Intel NUC6i7KYK onboard PC with an Intel Core i7-6770HQ quadcore CPU running at \SI{2.6/3.5}{\giga\hertz} and \SI{32}{\giga\byte} of RAM.
As primary environment perception sensor, a Velodyne Puck LITE lidar is deployed. It features a low weight of \SI{590}{\gram} and yields 300,000 range measurements per second in 16 horizontal scan lines at a vertical angle of $\SI{30}{\degree}$. Its maximum range is \SI{100}{\meter}.

In order to perceive visual tags in nearby shelf panels during flight, the MAV is equipped with two synchronized global shutter Point Grey Blackfly-S U3-51S5C-C color cameras with 5.0\,MP. The Computar M0814MP2 lens features an apex angle of \SI[product-units = single]{56,3 x 43,7}{\degree}. Each camera captures 3 frames per second. For detection of RFID tags, the MAV is also equipped with a ThingMagic M6e RFID reader with a SkyeTek SP-AN-04-UF-BB6LP antenna.

The low weight of the components (\SI{11.2}{\kilo\gram} take-off weight) and a battery capacity of 600\,Wh, yields a flight time of approximately \SI{20}{\minute} which allows to capture \SI{1}{\kilo\meter} of shelf. Since the batteries are hot-swappable, continuous operation can be performed with only minimal interruptions.

The system uses the robot operating system (ROS) as middleware on both the MAV and an additional ground control station. We show an overview of the system in \reffig{fig:system_diagram}.

%% file: environment_perception.tex
\subsection{3D Mapping}
\label{sec:3D_Mapping}
To localize the MAV within the environment, we build an allocentric map of the warehouse from measurements of the lidar. Fig.~\ref{fig:mapping} shows parts of the warehouse and the initial map.
We incorporate measurements of the IMU to account for motion of the sensor during acquisition.
Using an extended version of our lidar-based SLAM method described in~\cite{Droeschel2017104}, we first aggregate 3D scans in a local multiresolution grid map.
Local multiresolution maps correspond to the sensor measurement characteristics by having a high resolution close to the sensor and a coarser resolution farther away. 
For each grid cell, a local surface element (surfel) is estimated which summarizes the aggregated measurements in the cell's volume and captures the statistics of the points.
We recover the transformation between a newly acquired scan and the local map by matching surfels~\cite{Droeschel:ICRA2014}. 
Compared to point-based registration, considerably less elements are taken into account for registration, allowing for efficient registration of the extensive amount of measurements from the sensor. 

Registered 3D scans are added to the local map, replacing older measurements. 
Local mapping allows to track the robot in a local frame and provides a dense aggregation of measurements in the robot's vicinity. 

We construct an allocentric pose graph by aligning local multiresolution maps from different view poses which allows the robot to localize itself in an allocentric frame. Hence, local multiresolution maps from different view poses model nodes in a graph~$\mathcal{G} = ( \mathcal{V}, \mathcal{E} )$ that are connected by edges.
Edges model spatial constraints between nodes and result from aligning two local multiresolution maps by surfel-based registration.
The registration result~$x_i^j$ between a new node~$v_i$ and the previous node~$v_j$ constitutes an edge~$e_{ij} \in \mathcal{E}$.

Additionally, the current local map is registered towards a reference node in order to connect the current pose to the global pose graph and enable a straightforward optimization.
The reference node is the local map that is closest to the current MAV pose.
If the robot moved sufficiently far, we extend the pose graph by the current local map.

Furthermore, we include edges between the newly added local map and close-by local maps to obtain loop closure if the robot revisits previously mapped areas. 
Hence, we check for one new edge between the current reference~$v_{\text{ref}}$ and other nodes~$v_{\text{cmp}}$.
We determine a probability
\begin{equation*}
  p_\text{chk}(v_{\text{cmp}}) = \mathcal{N}\left(d(x_{\text{ref}}, x_{\text{cmp}}); 0, \sigma_d^2\right)
\end{equation*}
that depends on the linear distance~$d(x_{\text{ref}}, x_{\text{cmp}})$ between the view poses~$x_{\text{ref}}$ and~$x_{\text{cmp}}$.
According to~$p_\text{chk}(v)$, we draw a node~$v$ from the graph and determine a spatial constraint between the nodes using our surfel registration method.

From the spatial constraints, we infer the probability of the trajectory estimate given all relative pose observations
\begin{equation*}
 p( \mathcal{V} \mid \mathcal{E} ) \propto \prod_{e_{ij} \in \mathcal{E}} p( x_i^j \mid x_i, x_j ).
\end{equation*}
Each spatial constraint is a normally distributed estimate with mean and covariance determined by our probabilistic registration method.
This pose graph optimization is efficiently solved using g\textsuperscript{2}o~\cite{kuemmerle2011_g2o}, yielding maximum likelihood estimates of the view poses $x_i$.

We extend our mapping approach presented in~\cite{Droeschel2017104} to allow for efficient processing of Velodyne scans.
In contrast to the approach presented in our previous work, we do not aggregate multiple 3D scans using odometry information but register 
single 3D scans from the lidar sensor to the local multiresolution map---only using orientation information from the IMU and barometric height as prior for registration. 

\subsection{Lidar-based Localization}
\label{sec:Lidar_based_Localization}
Prior to autonomous operation, we acquire an initial map from a manual flight. We extend our SLAM system to serialize the graph-based structure of the allocentric map to gain persistent storage of the so-far acquired pose graph.

For autonomous operation during mission, the mapping system is initialized with the pose graph from the initial flight. By aligning the current local map to the pose graph, we gain a localization pose with 
respect to the initial map and the warehouse model. Although the pose graph (and the associated map) can be extended if the MAV traverses parts of the environment that where not covered by the initial flight, 
we choose the coverage volume of the initial flight to be larger than the MAV's workspace in the experiments since this is the envisaged operating mode during standard inventory missions.

While executing the mission, we localize towards the closest local map in the graph by registering the current local map with it.
Our approach allows to process the lidar scans in real-time.

\subsection{Tag Detection}
\label{sec:Tag_Detection}
We perceive the position of stock in the warehouse by means of visual fiducial markers (AprilTags) and RFID tags attached to storage boxes.
Perceived RFID tags, the current MAV position, signal strength, and direction of the detecting antenna are transmitted to the WMS for further processing, \eg assigning stock to storage units.
Regarding the fiducial markers, we use the implementation by Olson \etal~\cite{olson2011tags} and transform the camera-based relative tag pose into an allocentric frame via the known camera extrinsics and the estimated pose of the MAV. Likewise, these allocentric positions and the corresponding tag IDs are sent to the WMS for incorporation into the warehouse model. We use the tag family \texttt{36h11} as we experienced it to be very reliable.

\begin{figure}[t]
  \centering
  \includegraphics[trim=00mm 00mm 00mm 00mm,clip,height=0.295\linewidth]{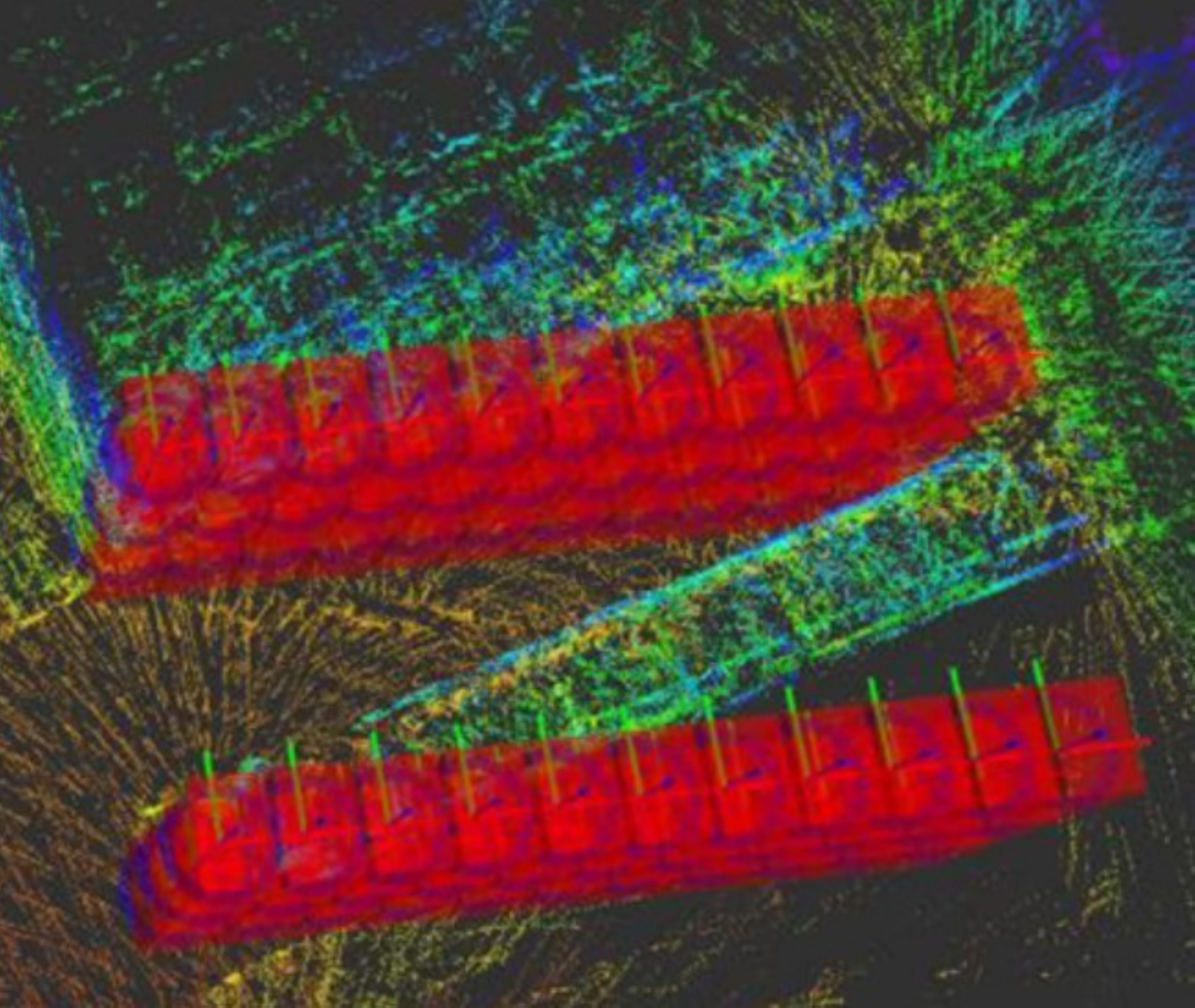}~
  \includegraphics[trim=00mm 00mm 00mm 00mm,clip,height=0.295\linewidth]{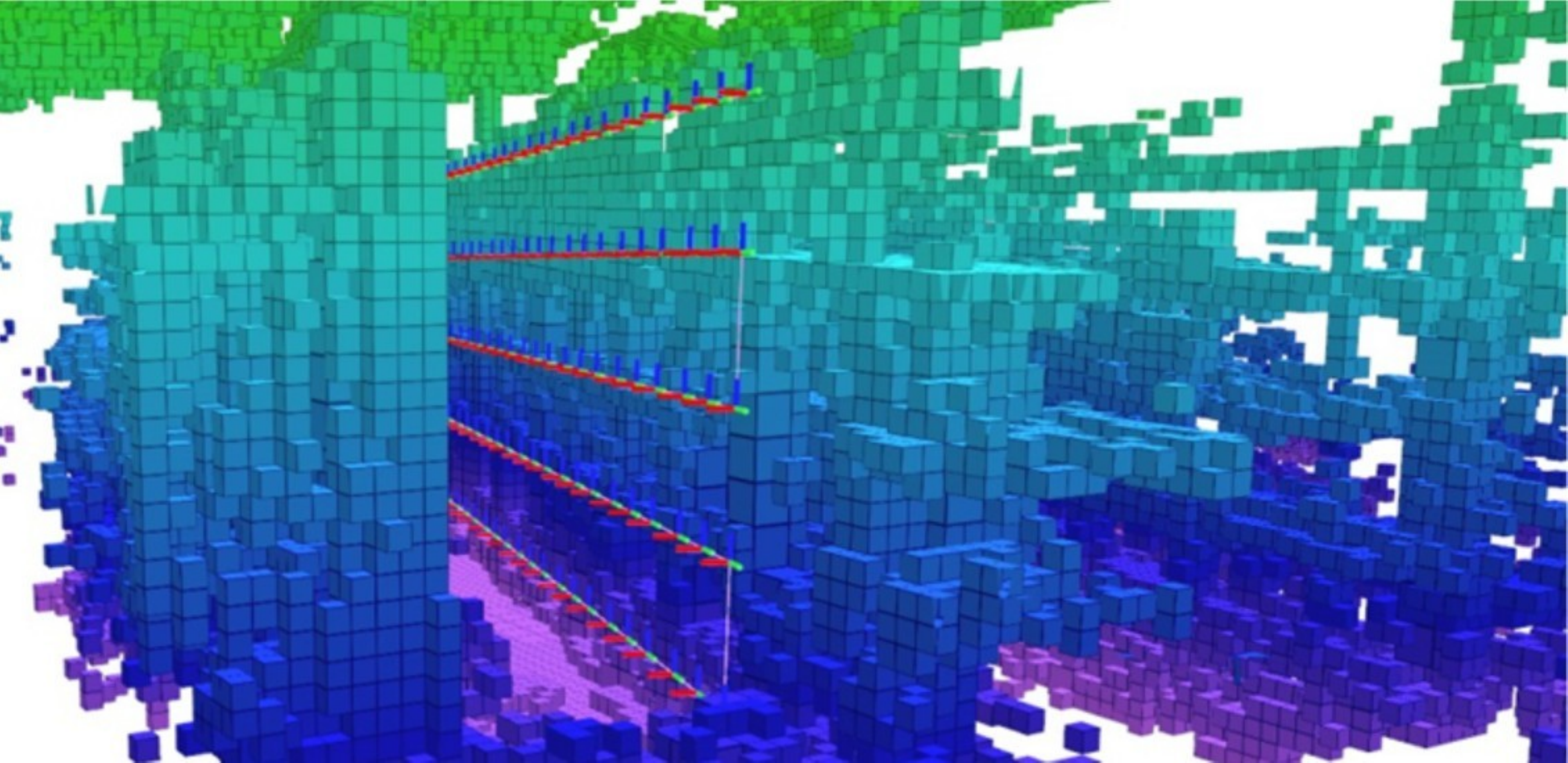}%
  \vspace{-1ex}
  \caption{Left: Registration of a semantic map (coordinates of storage units, geometric shelf model depicted in red) with a 3D laser map. Color encodes height. Right: Generated inventory mission (depicted by the coordinate axes) in the obstacle grid map.}
  \label{fig:registration}
  \vspace{-3ex}
\end{figure}

%% file: navigation_and_control.tex
Autonomous navigation is a key capability for automated stocktaking.
Operator assistance functions---or optionally fully autonomous operation---opens up the applicability of the system to a large group of end users who are not trained MAV pilots. Autonomy generates a direct interface between logistics personnel and the stocktaking system without the indirection of a professional pilot.
We implement a hierarchical navigation and control system that makes use of time scale separation between the layers. On the top layer of our navigation stack, global mission planning is executed once per mission. The next layer (allocentric path planning) is run in the order of seconds, while the lowest layer (model predictive trajectory planning) is executed every \SI{20}{\milli\second}.

\subsection{Mission Planning}
\label{sec:Mission_Planning}
For the connection to a WMS, we developed a tool that augments the laser-based maps described in Sec.~\ref{sec:Environment_Perception} with semantic information.
Fig.~\ref{fig:registration} shows the registration of the semantic warehouse model with the laser-based map. After a coarse manual alignment, we use the Iterative Closest Point Algorithm (ICP) to automatically register both maps.
This enables us to semantically describe an inventory mission and automatically derive shelf numbers and indices of storage places.
The WMS can specify missions covering whole shelves---with a coverage pattern shown in \reffig{fig:registration}---or single storage units to inspect.
Here, all common strategies for manual inventory like \eg sampling inventory with sequential probability ratio test (SPRT) can be utilized.
An ordered list of view poses is then sent to the MAV onboard computer for execution.
Before execution, this list is simplified to merge collinear path segments, \eg a number of storage units on the same height, to achieve a smooth sweeping motion along the shelves.

\begin{figure}[t]
  \centering
  \includegraphics[trim=0 150 0 50, clip, width=0.48\linewidth]{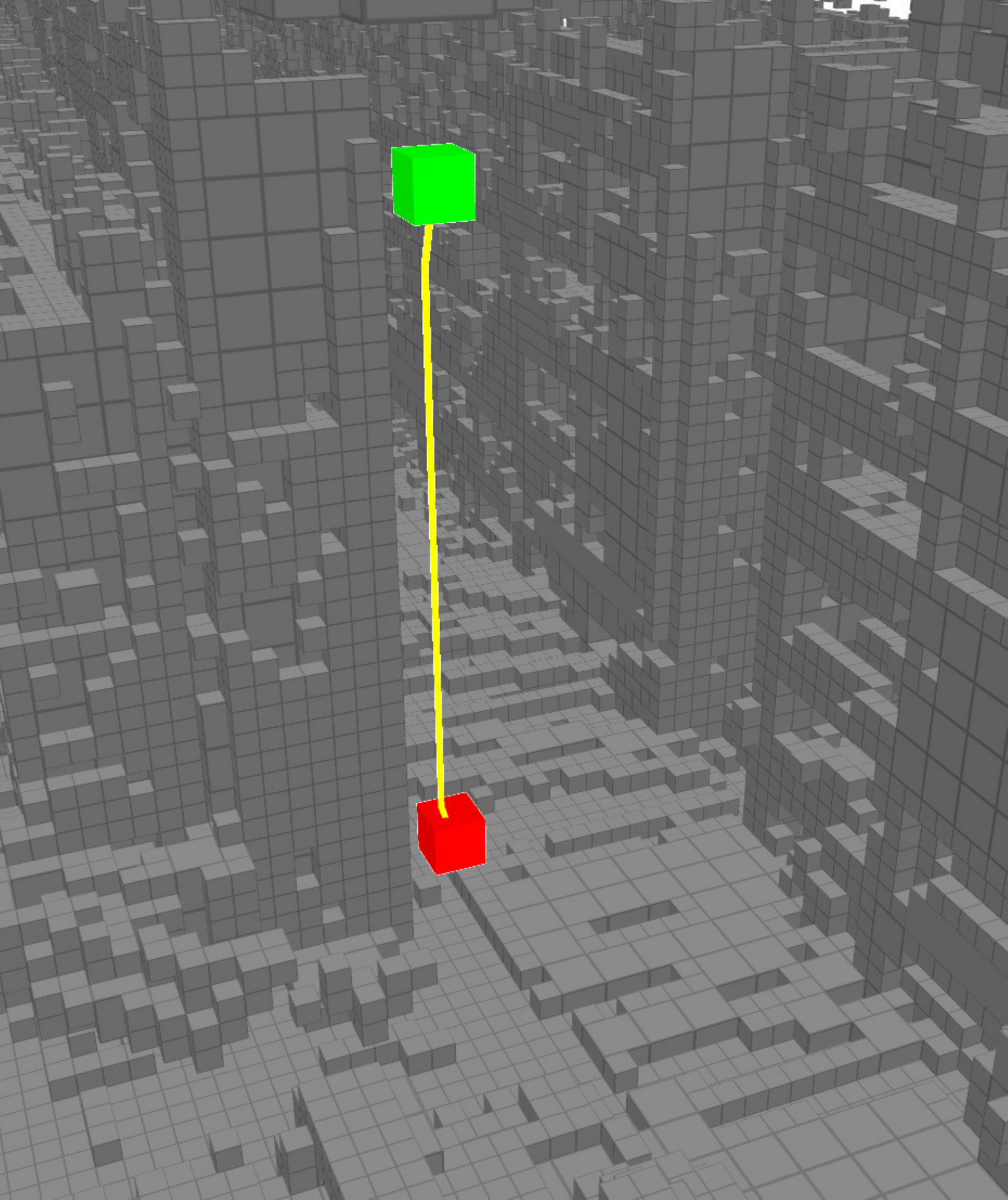}~
  \includegraphics[trim=0 150 0 50, clip, width=0.48\linewidth]{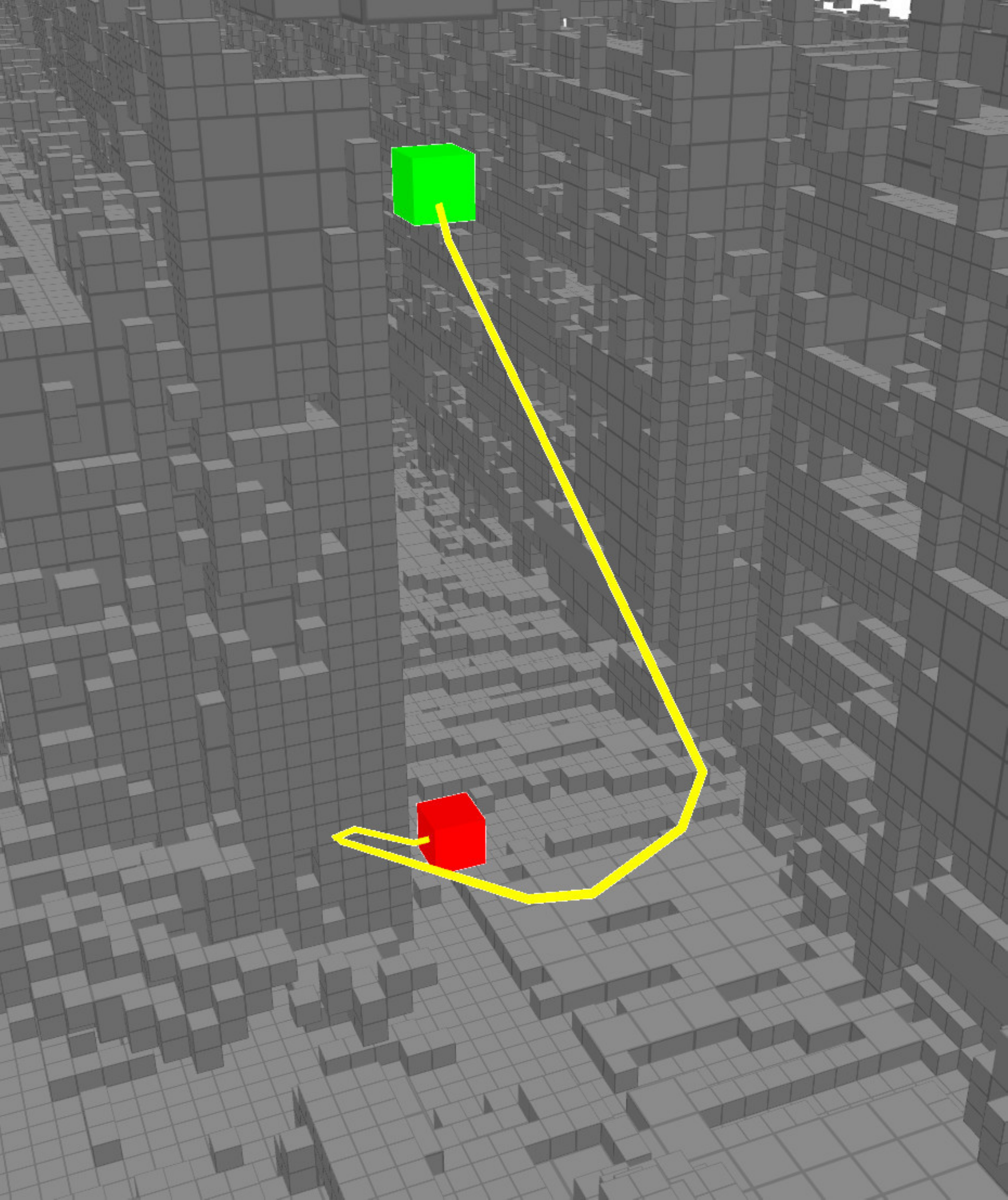}%
  \vspace{-1ex}
  \caption{Planning under visibility constraints. Left: Without visibility constraints the shortest path (yellow) from a start (green) to a target position (red) below solely descents in place. Right: With visibility constraints, the MAV has to move within the field of view of the lidar and consequently follows a longer descent path with an angle of \SI{15}{\degree}.}
  \label{fig:constrained_planning}
  \vspace{-2ex}
\end{figure}

\subsection{Path Planning}
\label{sec:Path_Planning}
The result of the mission planning is an ordered list of 4D-poses (x,y,z,$\theta$) in a discrete allocentric grid.
We connect these poses with an instance of A* planning and use the Ramer-Douglas-Peucker algorithm to cull superfluous nodes.
This is necessary to allow for the generation of longer and more continuous trajectories by our controller, described below.
During mission execution, the path is frequently replanned to compensate for path deviations of the MAV, either by inaccurate command execution, external disturbances or avoidance of obstacles.
Replanning takes place whenever a target pose is reached and the next pose from the mission plan is processed or at least every \SI{10}{\second} to correct deviations from the path.
Grid-based planning resembles the orthogonal structure of warehouses if the aisles are parallel to the planning grid axes.
The planning grid and the model are, therefore, aligned after the exploration flight.
Our approach, in contrast to sampling-based planners, has the advantage to follow the shelf-fronts well, without much postprocessing and trajectory smoothing.
In more generalized settings, the approach could benefit from any-angle planning, \eg Theta*~\cite{thetastar}, which we can omit here.

The onboard lidar does not cover a spherical field of view. To nevertheless allow for safe navigation in cluttered environments or in the presence of dynamic obstacles, we extended our planner with visibility-constrained planning.
With this extension, the planned MAV movements are restricted to directions in the field of view of the lidar, \ie \SI{15}{\degree} below and above the current horizontal plane.
To this end, we employ a grid with anisotropic voxels to reduce the ascent and descent angles from \SI{45}{\degree} in a grid with isotropic voxels to the opening angle of the sensor.
The resulting voxels have a height of $\tan (15^\circ) \approx \frac{1}{4}$ of the horizontal voxel size.
Furthermore, we remove edges connecting cells directly on top of each other, disallowing ascents and descents in place.
The direction of flight---discretized to the eight possible transitions in the plane---is introduced as a new planning dimension to penalize changes in the flight direction.
Angles of up to \SI{45}{\degree} are not penalized. Without this penalty, a zigzag motion to ascent or descent would be equal to larger straight glide paths in path costs, but would significantly slow down the MAV due to numerous stops to change direction.
\reffig{fig:constrained_planning} illustrates the resulting plans with and without visibility constraints.

\begin{figure}[t]
  \setlength{\figureheight}{0.41\linewidth}
  \centering
  \includegraphics[trim=0 0 0 150, clip, height=\figureheight]{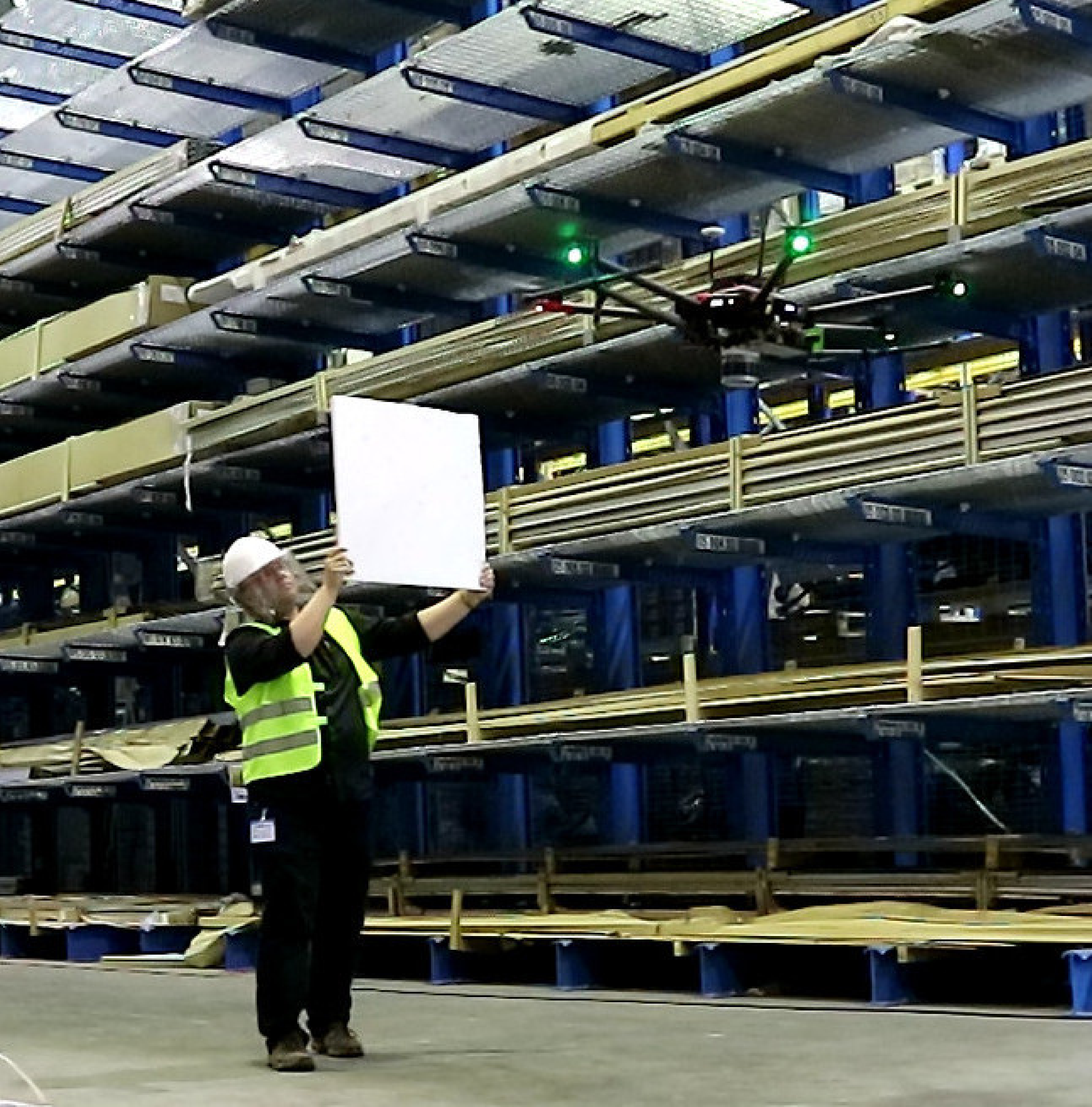}~
  \includegraphics[trim=0 0 0 150, clip, height=\figureheight]{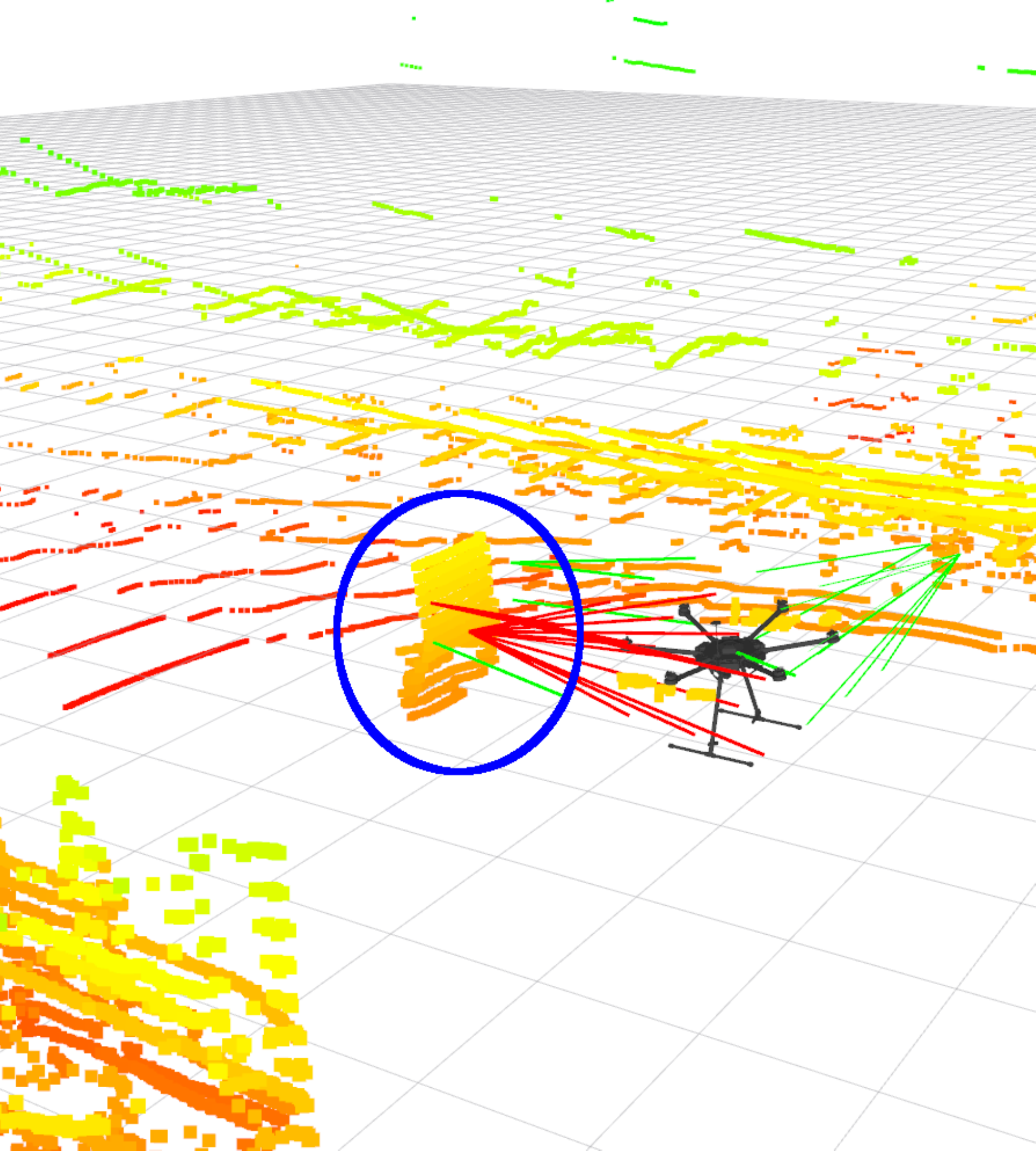}%
  \vspace{-1ex}
  \caption{Reactive obstacle avoidance with artificial potential fields. A person (circled blue in the laser map) approaches the MAV. The MAV is repelled by the artificial forces (red lines) and dodges the obstacle. Green lines depict the influence of obstacles in the passive avoidance distance.}
  \label{fig:hindernisvermeidung_screenshot}
  \vspace{-2ex}
\end{figure}

\subsection{Reactive Obstacle Avoidance}
\label{sec:Reactive_Obstacle_Avoidance}
We use reactive obstacle avoidance as a low-level safety layer complementing the deliberative path planning.
For our application, reactive obstacle avoidance has two important properties---compared to fast local planning~\cite{3dvfh+}, or optimization-based approaches~\cite{Israelsen2014}.
First, it has the ability to elude approaching dynamic obstacles, depicted in \reffig{fig:hindernisvermeidung_screenshot}. This might include leaving a hover position or even moving into the opposite direction of the commanded flight path.
Second, a hazard minimizing solution will always be found even if the distance constraints are violated.
Furthermore, reactive obstacle avoidance is computationally cheap and, consequently, can be executed with the lidar frequency of \SI{10}{\hertz}.
Our obstacle avoidance is based on~\cite{nieuwenhuisen2013isprs} but directly modifies the allocentric target waypoints from the global path planner instead of velocity commands to adapt to the new low-level trajectory controller.

\begin{figure}[t]
  \centering
  \includegraphics[width=0.32\linewidth]{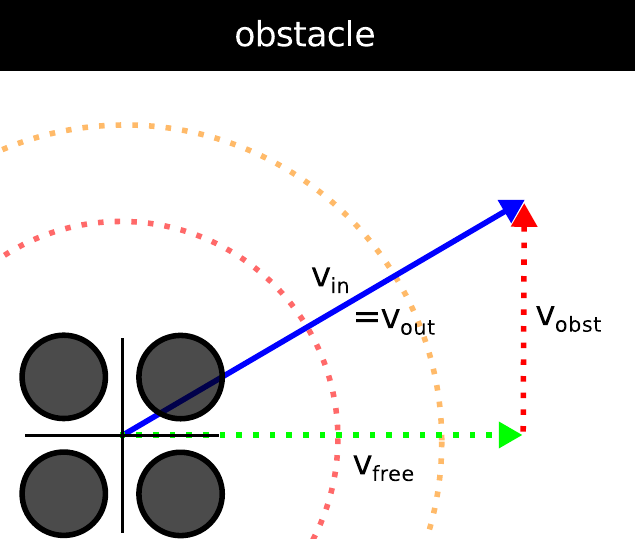}~
  \includegraphics[width=0.32\linewidth]{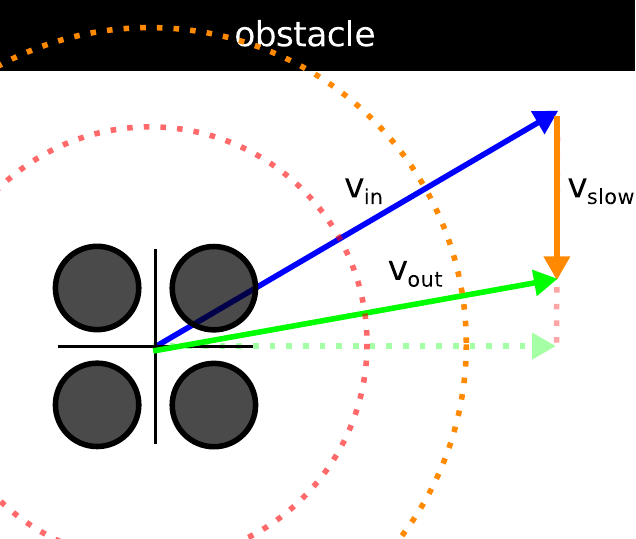}~
  \includegraphics[width=0.32\linewidth]{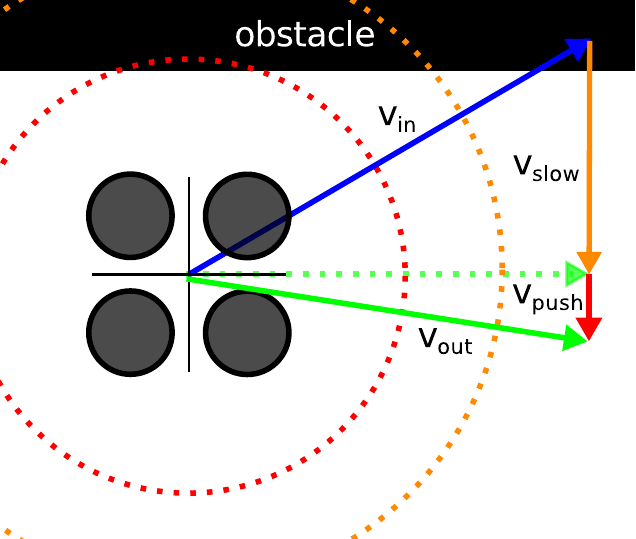}\\[2ex]

  \resizebox{0.8\linewidth}{!}{
  \input{obstacle_avoidance.pgf}
  }
  \vspace{-2ex}
  \caption{Reactive obstacle avoidance. Top-Left: The MAV velocity setpoint vector $v_{in}$ is split into the projection towards an obstacle $v_{obst}$ and the remainder $v_{free}$. If the MAV is not close to obstacles, the output velocity $v_{out}$ is equal to the setpoint. Top-Middle: When an obstacle is in the passive avoidance sphere (dotted orange), $v_{in}$ is reduced by $v_{slow} = -s_\mathrm{reduce} v_{obst}$. Top-Right: Obstacles in the active avoidance sphere (dotted red) induce an additional repulsive force resulting in the pushing velocity $v_{push}$ directing the MAV into free-space. For simplicity, we depict velocity vectors, the pose modification vectors $c_o$ and $f_o$ follow straightforward. Bottom: Scaling factors in relation to the obstacle distance.}
  \label{fig:obst-avoidance-vectors}
  \vspace{-2ex}
\end{figure}

We modified the basic algorithm to facilitate smoother flight in narrow spaces by adding two spheres of influence around the MAV, depicted in \reffig{fig:obst-avoidance-vectors}.
Obstacles in the passive avoidance sphere with radius $d_p$, cause a reduction of the MAV motion into the direction of the obstacles.
In the active avoidance sphere with radius $d_a$, obstacles exert artificial repulsive forces, increasing with proximity, that push the MAV away.
By dividing the obstacle avoidance into these two phases, we achieve a stable equilibrium distance between obstacles and MAV regardless of the MAV control inputs without influencing the motion into orthogonal directions in the passive sphere---\eg the MAV can follow an exploration pattern along a shelf even if the commanded pattern is too close to the shelf due to protruding goods.
In the warehouse, we set $d_a$ and $d_p$ to MAV radius plus \SI{1}{\meter} and \SI{2}{\meter}, respectively.

For simplicity of notation, all further calculations are depicted in an egocentric MAV frame to omit the localization transform matrices.
If both spheres are obstacle-free, we execute the commands from the planning layer unaltered.
Egocentric targets farther away than \SI{1}{\meter} are first normalized; shorter vectors are processed without prior normalization to avoid a speed up of the MAV while approaching an obstacle.
The new egocentric target position $t_{\textrm{new}}$ is calculated as
\begin{equation*}
  t_{\textrm{new}} = t_{\textrm{orig}} - c_o s_\mathrm{push} + f_o s_\mathrm{reduce}.
\end{equation*}
Here, $c_o$ is the projection of the current target $t_{\textrm{orig}}$ onto the direction of the obstacle, thus, the part of the command that steers the MAV closer to the obstacle.
The artificial force direction $f_o$ is a normalized vector pointing away from the obstacle.
The magnitudes of the slow down strength $s_\mathrm{push}$ and the push back strength $s_\mathrm{reduce}$---depicted in \reffig{fig:obst-avoidance-vectors}---are calculated as
\begin{equation*}
  s_\mathrm{push} = \frac{d_p + d_a - d}{d_p - d_a}, s_\mathrm{reduce} = \frac{d_a - d}{d_a}
\end{equation*}
with distance $d$ to the obstacle. Both results are clipped to the interval $[0,1]$ afterwards.

\subsection{Model Predictive Control}
\label{sec:Model_Predictive_Control}
Since higher layers assume a straight connection between waypoints (due to Ramer-Douglas-Peucker culling), flying on a straight trajectory is mandatory and overshoot is not permissible despite large turbulences caused by nearby obstacles. Also inventory of large warehouses with multiple kilometers of shelf requires fast flight to reduce the impact on the regular logistic processes.
We tackle this problem by employing time-optimal trajectory generation and online replanning with \SI{50}{\hertz} for low-level control. We use an extended version of the method described in~\cite{beul2017icuas}. Planning is based on a simple dynamic model of the MAV with three-dimensional jerk $j$ as only input. 
The method plans smooth, time-optimal trajectories from the current 9-dimensional allocentric MAV state
\begin{equation*}
\hspace{2.0em}
\bm{x} =
\begin{pmatrix}
p_{x} & p_{y} & p_{z}\\
v_{x} & v_{y} & v_{z}\\
a_{x} & a_{y} & a_{z}
\end{pmatrix}
\end{equation*}
to the corresponding 9-dimensional target state by analytically solving a system of 21 differential equations
\begin{align*}
    p_{n} &= p_{n-1} + \int_{t_{n-1}}^{t_{n}}{v_{n}}\,\mathrm{d}t,\\
    v_{n} &= v_{n-1} + \int_{t_{n-1}}^{t_{n}}{a_{n}}\,\mathrm{d}t, \qquad n = \{1;\dots;7\}\\
    a_{n} &= a_{n-1} + \int_{t_{n-1}}^{t_{n}}{j_{n}}\,\mathrm{d}t
\end{align*}
per axis (x,y,z).
Generated trajectories consist of up to $n = 7$ phases of constant jerk input, resulting in a bang-singular-bang trajectory.
Individual axes are coupled by synchronizing the total time of the entire trajectory.
The trajectories respect per-axis constraints on minimum and maximum velocity, acceleration and jerk.

\begin{figure}
    \centering
    \begin{tikzpicture}
        \node[inner sep = 0,anchor=north west] at (0,0) {\includegraphics[trim=010mm 000mm 000mm 000mm,clip,width=1.0\columnwidth]{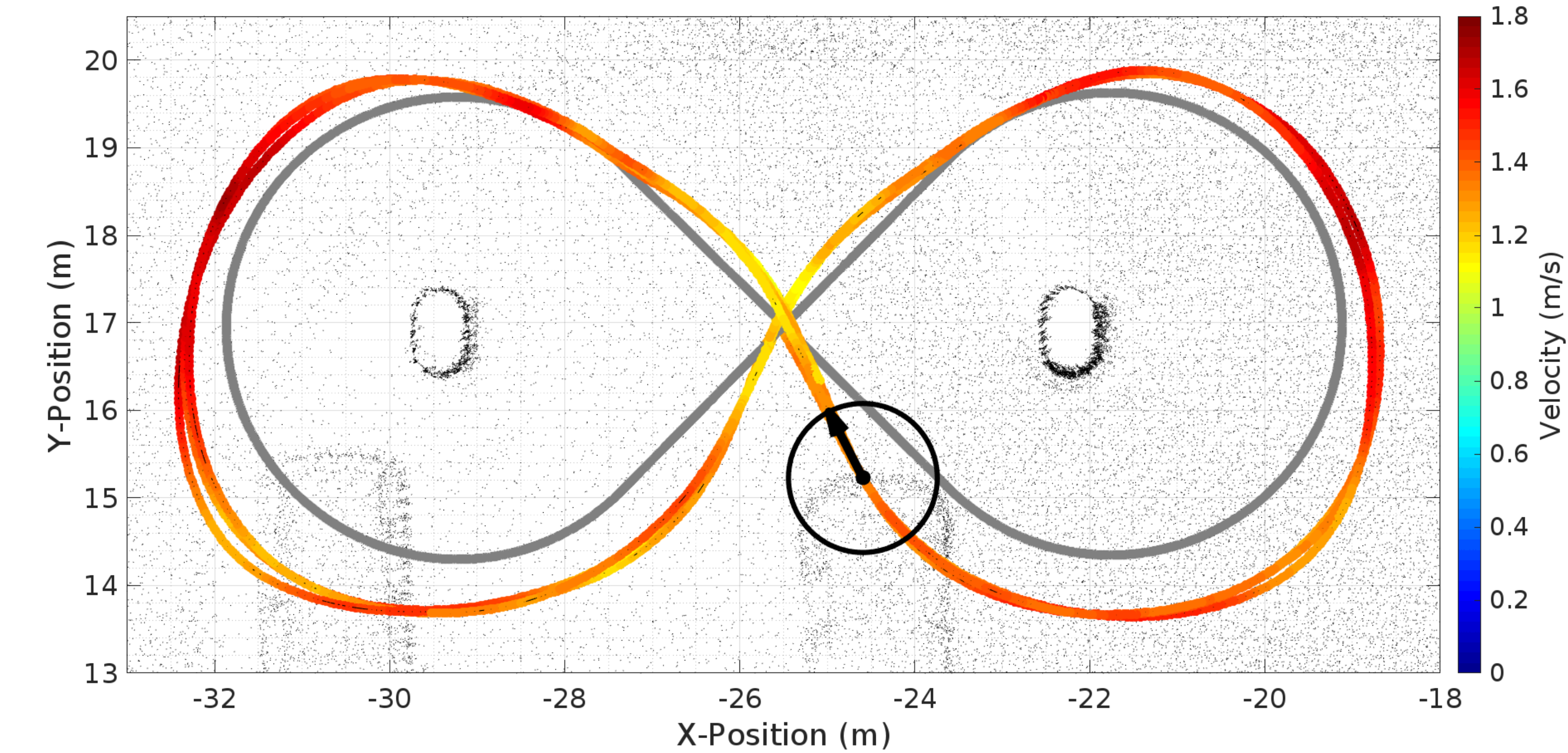}};
        \node[inner sep = 0,anchor=north west] at (0.43,-0.1) {\includegraphics[trim=000mm 000mm 000mm 000mm,clip,width=0.35\columnwidth]{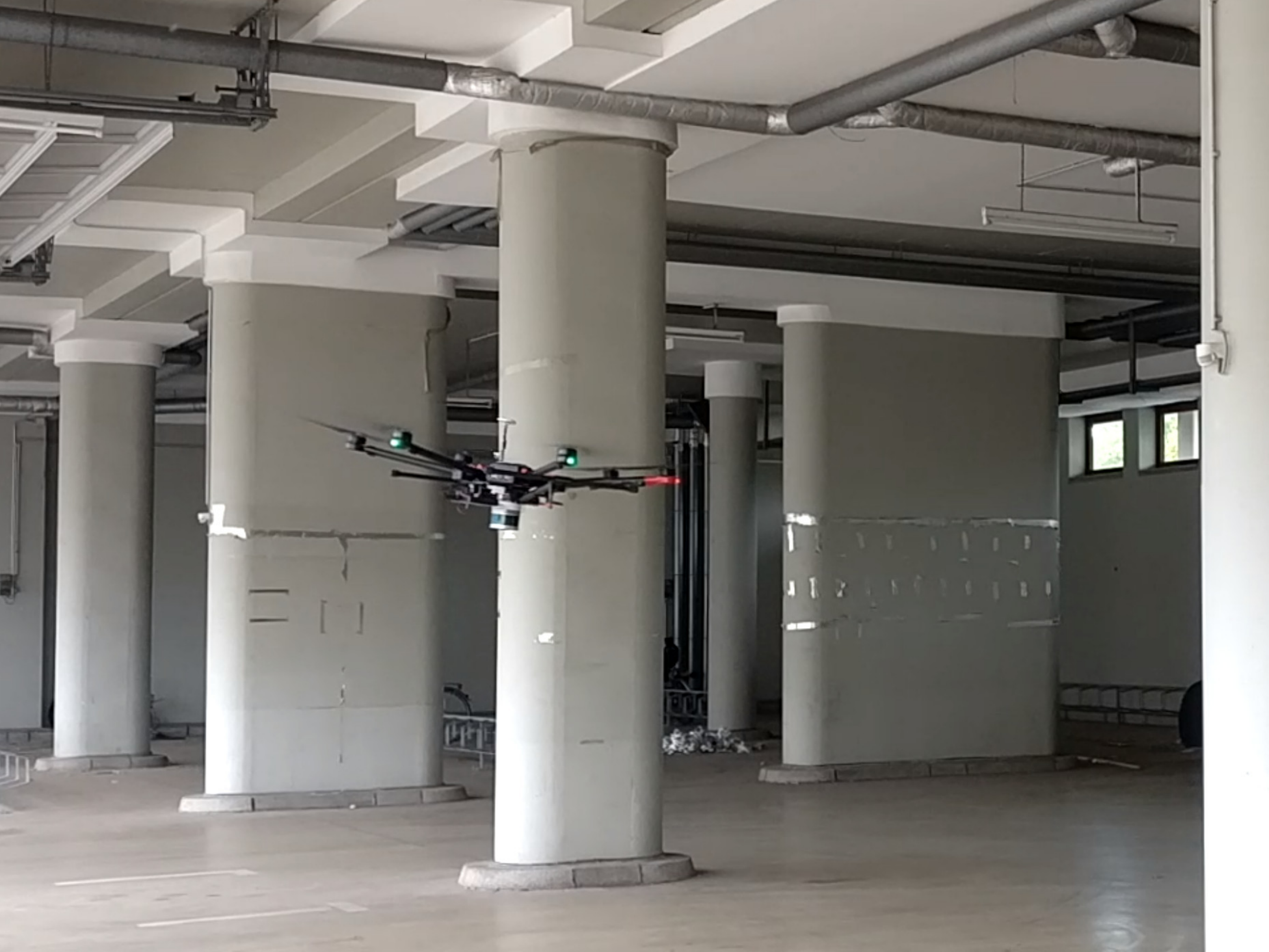}};
    \end{tikzpicture}
    \vspace{-2em}
    \caption{The MAV continuously flies in a figure eight around pillars in a parking garage. All perception and computation is done onboard. Velocities exceeding \SI{1.7}{\meter\per\second} in the vicinity of obstacles require robust methods for state estimation and control. The size of the ring represents the actual MAV size. The arrow depicts the flight direction.}
    \label{fig:trajectory_garage_image}
    \vspace{-1ex}
\end{figure}

With the ability to predict the target state, trajectories end in an optimal interception point when the waypoint is non-stationary like shown in \reffig{fig:trajectory_garage_image}.
Since our method is very fast, we use it in closed loop and send smooth pitch $\theta$, roll $\phi$ and climb rates $v_{z}$ to the DJI flight control.

We assume the yaw to be decoupled from the translatory axes and use simple proportional control for the yaw. The yaw rate setpoint $\dot\Psi_{setp} = K_{p} \cdot (\Psi_{setp} - \Psi)$ with proportional gain $K_{p}$ is sent to the MAV.

In comparison to approaches that utilize a complex motion model like Kamel \etal~\cite{kamelmpc2016}, our approach is very fast and the model does not need (often abstract) parameters.
In contrast to complex models, approaches like Mueller \etal~\cite{Mueller2013_2} use a simple motion model and are comparably fast. The generated trajectories however are not time-optimal.
Simple PID-control is also not suitable, since overshoot is not permissible in the close corridors. Thus, the controller would have to be parameterized very conservative which would result in slow MAV movement.

Please note that we use the same parameters for the controller as in~\cite{ecmr2017_c1} in which we employed the method on a DJI Matrice 100 that weighs only a quarter of the MAV used here with a corresponding bounding box volume ratio of 1:12. This shows the independence of our approach regarding model parameters.

\begin{figure}[t]
  \centering
  \begin{tikzpicture}
      \node[inner sep = 0,anchor=north west] at (0,0) {\includegraphics[trim=000mm 000mm 000mm 000mm,clip,width=1.0\columnwidth]{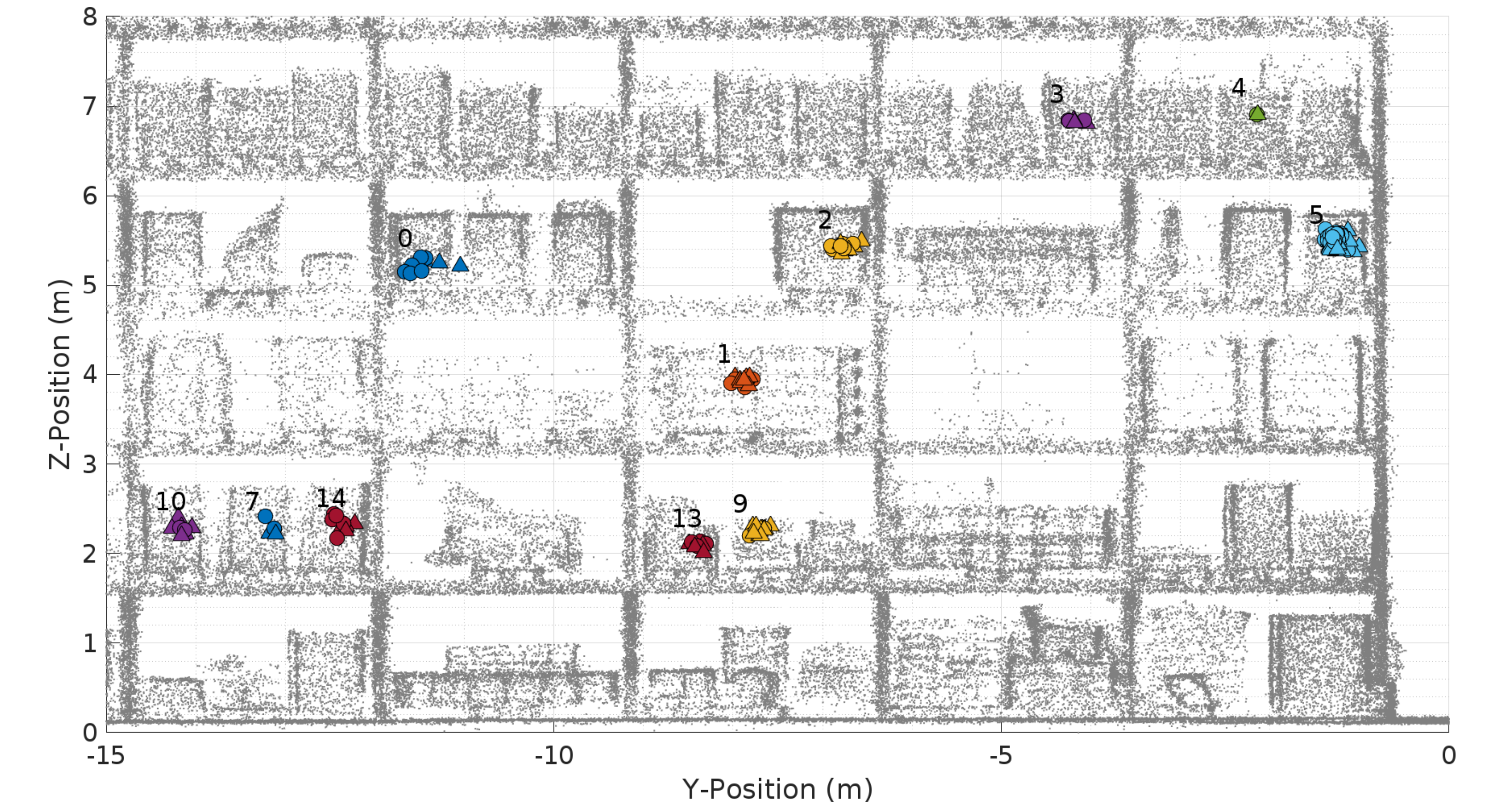}};
      \node[inner sep = 0,anchor=north west] at (6.13,-1.70) {\includegraphics[trim=110mm 010mm 100mm 145mm,clip,width=0.25\columnwidth]{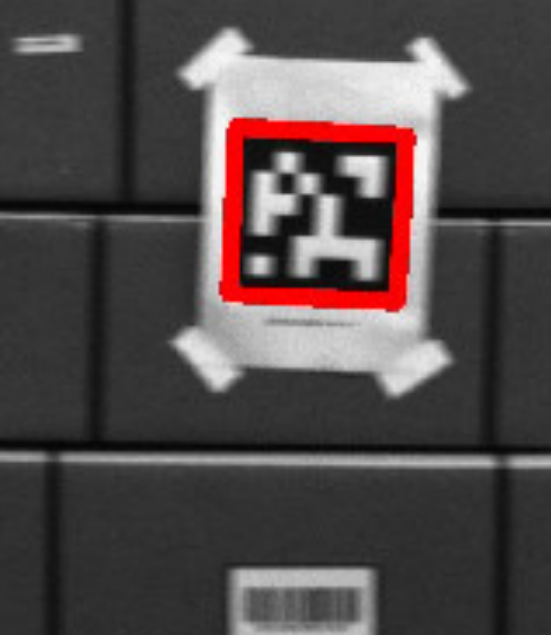}};
  \end{tikzpicture}
  \vspace{-2em}
  \caption{AprilTag detection. With two cameras directed to each side of the aisle we detect AprilTags attached to the stock. The clusters of colored markers show the estimated positions of the detected tags in a subsection of an aisle during two consecutive flights. See \reffig{fig:trajectory_image} for the corresponding trajectories. Detections from the first flight are marked with circles; detections from the second flight are marked with triangles. Different colors correspond to different tag IDs. In the bottom right corner, a detection of ID 14 is shown. One can see the motion-blur induced by the relative motion between AprilTag and MAV. We report statistics in \reftab{tab:results}.}
  \label{fig:detections}
  \vspace{-1ex}
\end{figure}

\begin{figure*}[t]
  \centering \footnotesize
  \fboxsep0mm
  a)\includegraphics[trim=028mm 000mm 035mm 000mm,clip,height=0.31\textwidth]{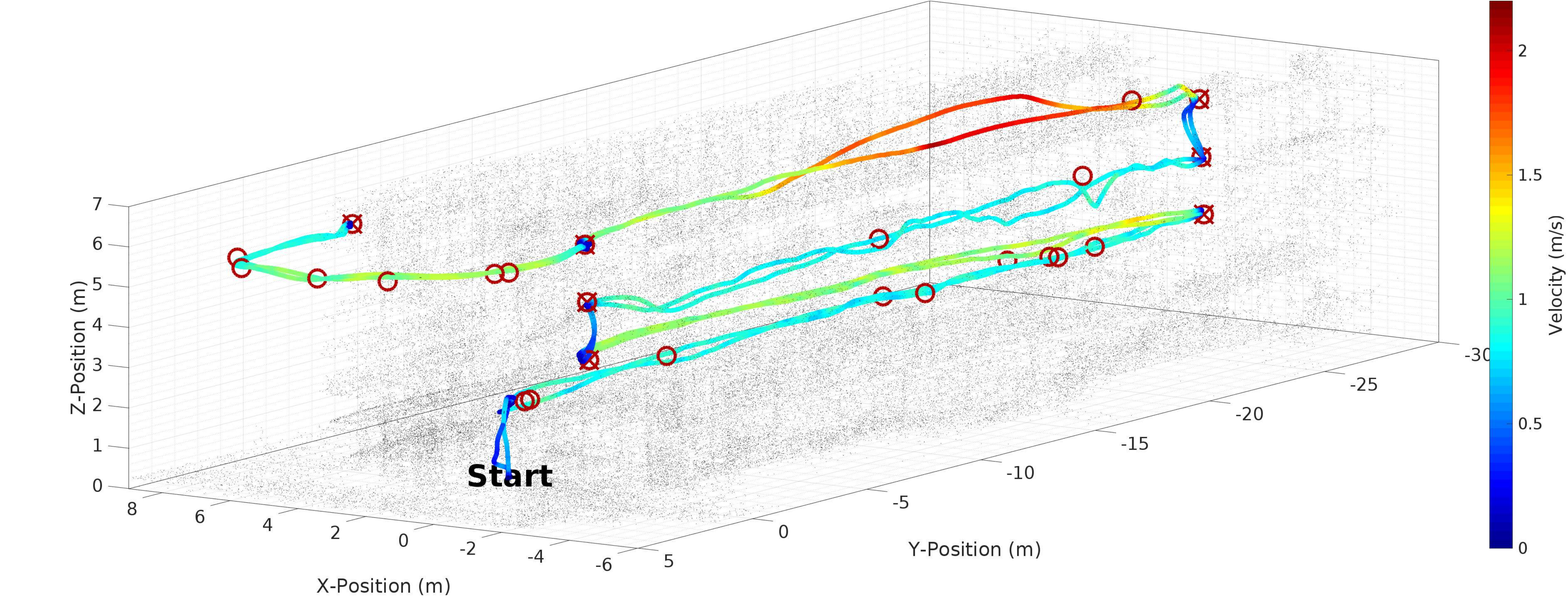}%
  b)\includegraphics[trim=255mm 000mm 220mm 000mm,clip,height=0.31\textwidth]{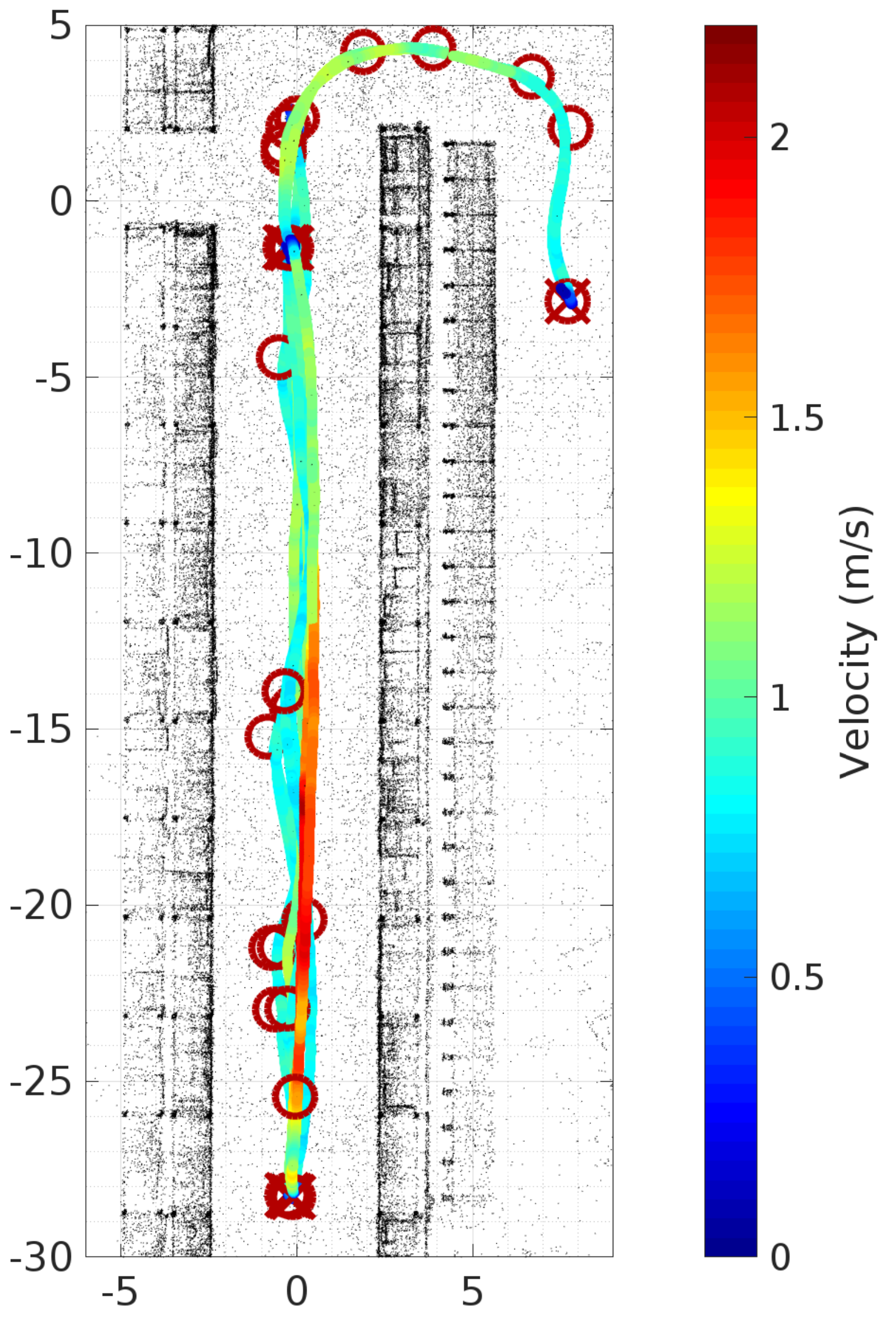}%
  \vspace{-1ex}
  \caption{Visualization of two consecutive flights in a warehouse. a) Side view, b) Top view. Despite flying in close proximity to obstacles, the MAV reaches velocities up to \SI{2.1}{\meter\per\second}. Waypoints are precisely reached without overshoot. It can be seen that the flight behavior is repeatable and that aisle changes are possible. View poses are marked with a red crossed ring. Via points that are inserted by the A* planner are marked with a red ring. Except for the manual start, the whole flight was fully autonomous.}
  \label{fig:trajectory_image}
  \vspace{-1ex}
\end{figure*}

%% file: obstacle_avoidance.pgf
\begin{tikzpicture}[auto]
\tikzset{content_node/.append style={minimum size=2em,minimum height=3em,draw,align=center,rounded corners,scale=0.65}}
\tikzset{label_node/.append style={scale=0.65,align=center}}
\tikzset{group_node/.append style={align=center,rounded corners,inner sep=1em,thick}}

\definecolor{red}{rgb}     {0.7,0.0,0.0}
\definecolor{green}{rgb}   {0.0,0.7,0.0}
\definecolor{blue}{rgb}    {0.0,0.0,0.7}
\definecolor{grey}{rgb}    {0.5,0.5,0.5}

\node(critical_distance)[label_node] at (1.,-0.4) {Critical\\distance};
\node(safety_distance)[label_node] at (3.,-0.4) {Active avoidance\\sphere radius};
\node(freespace_distance)[label_node] at (5,-0.4) {Passive avoidance\\sphere radius};
\node(x_label)[label_node] at (6.5,-0.4) {Distance\\to obstacle};
\node(y_label)[label_node] at (-0.2,0.57) {\rotatebox{90}{Strength}};
\node(y_tic)[label_node] at (-0.2,1.2) {1};

\draw[->, thick] (-0.3,0) -- (0,0 -| x_label.east);
\draw[->, thick] (0,-0.3) -- ([yshift=0.5em]y_tic.north -| 0,0);

\draw[-, thick,blue,dashed] (y_tic -| 0.,0.) -- (y_tic -| safety_distance) -- (0,0 -| freespace_distance) -- ( 0,0 -| x_label.east);
\draw[-, thick,red] (y_tic -| 0.,0.) -- (y_tic -| critical_distance) -- ( 0,0 -| safety_distance) -- (0.,0. -| x_label.east);

\draw[-, thick,red] ([yshift=0.2em]y_label -| x_label) -- ([yshift=0.2em,xshift=-2em]y_label -| x_label);
\draw[-, thick,blue,dashed] ([yshift=-0.8em]y_label -| x_label) -- ([yshift=-0.8em,xshift=-2em]y_label -| x_label);
\node(s_push)[label_node] at ([yshift=0.2em,xshift=1em]y_label -| x_label) {$s_{\mathrm{push}}$};
\node(s_slow)[label_node] at ([yshift=-0.8em,xshift=1.2em]y_label -| x_label) {$s_{\mathrm{reduce}}$};

\draw[-, thick] (0.,-0.1 -| critical_distance) -- (0,0.1 -| critical_distance);
\draw[-, thick] (0.,-0.1 -| safety_distance) -- (0,0.1 -| safety_distance);
\draw[-, thick] (0.,-0.1 -| freespace_distance) -- (0,0.1 -| freespace_distance);
\draw[-, thick] (-0.1,0 |- y_tic) -- (0.1,0 |- y_tic);

%

\end{tikzpicture}

%% file: evaluation.tex
We evaluate our system in indoor and outdoor scenarios, including an inventory mission in an active warehouse. A video showing autonomous mission execution and reactive obstacle avoidance can be found on our website\footnote{\scriptsize{\url{http://www.ais.uni-bonn.de/videos/IROS_2018_InventAIRy}}}. Here, we also publish recorded datasets, tools, and parts of our pipeline.

First, we test the robustness of the localization and control pipeline with an experiment that involves fast flight between alternating waypoints in an obstacle free courtyard over a distance of \SI{25}{\meter}.
The localization in an allocentric map of the courtyard and state estimation of the MAV was solely based on the onboard 3D lidar and the IMU; no GNSS feedback was used.
Between the acceleration and deceleration phases of the flight, the MAV reached a maximum velocity of \SI{7.8}{\meter\per\second}, measured by the onboard DJI GPS -- considered as ground truth.
The laser localization was running at \SI{20}{\hertz} to account for the large velocities. It was able to robustly track the MAV pose during the whole flight.
Despite strong wind, the maximum deviation from the straight line connection between both waypoints was only \SI{49}{\centi\meter} during all 11 alternations. The maximum overshoot recorded during the experiment was \SI{1.2}{\meter}.

In a second experiment, the MAV flew a figure eight around two pillars in a garage to additionally test the influence of turbulences close to structures and the ground.
Due to the high accelerations of approximately \SI{0.85}{\meter\per\second\squared} in the curved segments of the trajectory, the maximum velocity in these runs was reduced to yield a feasible, collision-free trajectory.
Still, the MAV reached velocities up to \SI{1.75}{\meter\per\second} in this indoor environment.
The laser localization tracked the MAV pose with \SI{10}{\hertz} and was able to keep the MAV localized in the map of the garage at all times.
\reffig{fig:trajectory_garage_image} shows the resulting trajectory in the map of the garage. It can be seen that our method yields robust repeatability in four consecutive flights despite turbulences.
Nevertheless, it can be seen that the MAV spirals out of the curved segments as it cannot accurately track the moving waypoint.

In a third experiment, our integrated system, including laser-based localization, planned navigation, obstacle avoidance, and acquisition of information about stock positions, was demonstrated in a warehouse with a building area of \SI[product-units = single]{100 x 60}{\meter} with \SI{1.3}{\kilo\meter} shelf (approximately \SI{12000}{\meter\squared} storage front).
As described in \refsec{sec:3D_Mapping}, we built an initial laser-based SLAM map of the environment with a manual flight, shown in \reffig{fig:mapping}.
This map is aligned with the semantic map containing storage units from the WMS.
For the demonstration of autonomous inventory, a mission containing the complete inventory of one shelf row and the inspection of a single storage unit in another row was specified in the WMS.
The MAV executed this mission autonomously multiple times while avoiding static obstacles, \eg the shelves and stock protuding from the shelves.
In \reffig{fig:trajectory_image} we visualize the trajectory of two consecutive flights in the warehouse. The MAV reaches velocities up to \SI{2.1}{\meter\per\second}.
Although faster flight is possible (as shown in the previous experiments), we used the ability of our MPC to limit the maximum velocity a) to account for the acceleration/deceleration distance needed by the MAV and b) to reduce motion blur in the cameras (see \reffig{fig:detections}):

The closed loop dynamics of MAV and MPC dictate the dynamic behavior of the system. Even under the assumption that the MAV is able to perfectly track the trajectories generated by the MPC and without any perception- or communication delay, an acceleration/deceleration distance of \SI{12.1}{\meter} is necessary with a maximum velocity of \SI{7.8}{\meter\per\second} (with parameters $a_{max} = \SI{3.5}{\meter\per\second\squared}, j_{max} = \SI{4.0}{\meter\per\second\cubed}$).

Furthermore, due to the artificial lighting in the warehouse, the camera exposure time had to be set to at least \SI{4}{\milli\second} for acceptable image quality. The used AprilTags have an edge length of \SI{16}{\centi\meter} that results in a patch size of \SI[product-units = single]{2 x 2}{\centi\meter}. Thus, the Nyquist frequency limits the relative velocity to \SI{10}{\meter\per\second} under ideal conditions. This velocity, however, would require special signal reconstruction techniques to preprocess the image for the AprilTag detector.
Also roll, pitch, and yaw motion superimpose the linear MAV velocity and generate relative motion between tag and MAV.
High-frequency vibrations generated by the propellers provoke additional blur. Therefore, we conservatively constrained the linear velocity in favor of robust detections in the warehouse experiment.

In contrast to the visual detection pipeline, the RFID reader did not limit the inventory speed since it is able to read up to 750 tags per second. We throttled the speed to 20 reads/second which was enough for our experiments and allowed for a higher detection range.

Every view pose is reached with a mean deviation of only \SI{9.65}{\centi\meter} respectively \SI{5.78}{\centi\meter} in both flights.
As no dynamic obstacles above the MAV were to be expected in this demonstration, we neglected the planning with visibility constraints in favor of faster mission execution.

During both flights, AprilTags on the sides of the aisle and RFID tags of the specified shelf row were captured and sent to the WMS.
\reffig{fig:detections} and \reftab{tab:results} show the result of the two flights. It can be seen that except for Tag 6, and 11, all tags are reliably detected (Tags 8 and 12 were not used). Our method was unable to detect Tag 11 due to a shadow that partially covered the tag on a disadvantageously positioned stock. Tag 6 was not attached properly and was flipped by turbulent air from the MAV. Not a single false positive detection happened during the experiment. It can be seen that only minimal scattering occurs and thus the relative detection error is small.

\begin{table}[t]
  \caption{Statistics of AprilTag detections for two flights. }
\small
 \setlength{\tabcolsep}{1.5mm}
  \vspace{-1ex}
\centering
  \begin{tabular}{@{}l@{\hspace{1pt}}ccccccccccc@{}}
  \toprule
  Tag ID               & 0              & 1                      & 2                       & 3                      & 4                       & 5                      & 7                      & 9                      & 10                     & 13                     & 14\\
  \midrule
  $n_{1}$              & 3              & 7                      & 6                       & 3                      & 1                       & 64                     & 2                      & 10                     & 6                      & 4                      & 6\\
  $n_{2}$              & 7              & 7                      & 6                       & 3                      & 1                       & 41                     & 3                      & 5                      & 3                      & 4                      & 5\\
  \midrule
  $\sigma_{1}$    & 10.4           & 3.2 & 4.7  & 2.7 & -                       & 4.7 & -                      & 3.3 & 4.2 & 3.3 & \phantom{0}3.9\\
  $\sigma_{2}$    & \phantom{0}3.9 & 4.3 & 5.0  & 3.4 & -                       & 3.8 & 3.3 & 3.5 & 2.1 & 2.8 & \phantom{0}4.8\\
  $|\mu_{1-2}|$  & 28.6           & 2.9 & 8.9  & 5.5 & 3.3  & 2.1 & 1.2 & 3.2 & 1.0 & 2.1 & 10.7\\
  \bottomrule    
\end{tabular}
\vspace{0.5ex}
\footnotesize 
   \rule[-.3\baselineskip]{0pt}{\baselineskip} \hspace*{-2ex} $n_i$ is the number of detections per flight, $\sigma_i$ the deviation of the\\ 
    detections in cm. $|\mu_{1-2}|$ is the distance of the means $\mu_i$ in cm.
  \label{tab:results}
\end{table}

After the executed inventory mission, the MAV hovered at a height of \SI{2}{\meter} above the ground.
A person approached the MAV, which avoided the dynamic obstacle by means of our reactive obstacle avoidance, shown in \reffig{fig:hindernisvermeidung_screenshot}.
Furthermore, a person stepped into the way of the MAV while it approached a waypoint.
The MAV stopped at a safe distance in all cases.

As shown in the experiments, the limiting factor for faster inventory is motion blur in the cameras caused by the large exposure time due to bad lighting conditions. In future work, we want to oppose this bottleneck either by illuminating the scene ourselves (by using a flash on the MAV) or by using special equipment like \eg event based cameras.

In the current setup, the MAV continuously records images with \SI{3}{\hertz}. The AprilTags however only cover less than \SI{3}{\percent} of the area (\SI{0.0256}{\meter\squared} tag size vs. \SI{0.96}{\meter\squared} storage unit front). A more targeted strategy would reduce the generated data.
Furthermore, we also plan to extend our vision pipeline to not only detect AprilTags, but also other visual indicators like, \eg barcodes, QR codes, and human readable text, commonly found on stock. This would further enhance the versatility of the system.
We also plan to integrate multiple MAVs into the mission planner for simultaneous inventory to speed up the process even more.

One might also think of eliminating the first manual flight in favor of an automated SLAM process, but that a manual flight is more robust in this crucial map building phase. Furthermore, in comparison to the service life of such a system, the map building phase only causes a small fixed effort, since the map is reusable. After the manual flight, the operators can check the map for possible artifacts and misregistrations.

%% file: conclusion.tex
In this paper, we presented an MAV that is capable of fast autonomous indoor and outdoor flight without the aid of external infrastructure, solely relying on an omnidirectional laser scanner for localization.
We approached this challenge by employing fast 6D lidar based localization in 3D maps in combination with time-optimal model predictive control. Due to the fast runtime of our methods, the MAV motion can be tracked and controlled even under high velocities and accelerations.
Our ROS-based mapping and navigation pipeline allows for fully autonomous flight even in GNSS-denied environments.

Ample onboard processing power in combination with a high bandwidth ground connection and long battery life leads to a system that is suitable to be deployed, \eg in warehouses for extensive stocktaking applications.
We demonstrated the system robustness in multiple experiments where the only manual interactions were the starting and landing phases.